%% file: main.tex
\documentclass[runningheads]{llncs}

\usepackage{eccv}

\usepackage{eccvabbrv}

\usepackage{graphicx}
\usepackage{booktabs}
\usepackage{soul}
\usepackage{xcolor}

\usepackage[accsupp]{axessibility}  
\usepackage{wrapfig}
\usepackage{caption}

\usepackage[pagebackref,breaklinks,colorlinks,citecolor=eccvblue]{hyperref}

\usepackage{orcidlink}

\usepackage{tikz}
\usetikzlibrary{shapes.geometric, arrows.meta, positioning, fit, decorations.pathreplacing, shadows.blur, patterns, backgrounds, shadows}
\usepackage{fontawesome5} 
\usepackage{pifont} 
\usepackage{xcolor}
\usepackage{colortbl}

\input{sec/preamble}

\begin{document}

\title{Learning to Rank Caption Chains for Video-Text Alignment} 

\titlerunning{Learning to Rank Caption Chains for Video-Text Alignment}

\author{Ansel Blume\thanks{Work performed as an intern at Amazon Prime Video.}\inst{1} \and Burak Uzkent\inst{2} \and Shalini Chaudhuri\inst{2} \and Garin Kessler\inst{2}}

\authorrunning{A. Blume et al.}

\institute{
    University of Illinois Urbana-Champaign \\
    \email{blume5@illinois.edu}
    \and
    Amazon Prime Video
}

\maketitle

\input{sec/abstract}

\input{sec/intro}

\input{sec/related_work}
\input{sec/method}

\input{sec/experiments}
\input{sec/conclusion}

\bibliographystyle{splncs04}
\bibliography{references}
\end{document}


\title{Appendix}

\author{}
\institute{}

\maketitle
\thispagestyle{empty}
\appendix


\section{Mitigating Bias in Caption Evaluations}
\subsection{Experimenting with Another Judge}
As noted in the main text, we experiment with GPT-OSS 20B as a judge to ensure that the evaluations are not biased due to our using Claude 3.7 Sonnet to both generate the caption chains and evaluate all of the models. \cref{tab:captioning-gpt_oss} shows the complete results. While GPT-OSS has lower baseline scores for all but fluency, the trends we observed with the Claude evaluations \textbf{generally hold} with GPT-OSS as a judge, \textbf{confirming that our findings are not an artifact of judge selection}.

\begin{table*}[h!]
    \small
    \centering
    \caption{Captioning results on PE-Video and MSR-VTT splits of \cds using GPT-OSS as a judge to ensure that evaluations are not biased by Claude. Captions are evaluated by Relevance (\textit{Rel.}), Descriptiveness (\textit{Descr.}), Temporal Consistency (\textit{Temp C.}), and Fluency (\textit{Flu.}). \textbf{Our ranking method achieves the best or tied-best results across nearly all metrics and datasets.}}
    \begin{tblr}{
        colspec = {l *{8}{Q[c]}},
        colsep = 4pt,
        rowsep = 1pt,
        row{1} = {font=\bfseries, bg=gray!8},
        row{2} = {font=\itshape, bg=gray!8},
        row{3} = {bg=blue!8, font=\itshape, rowsep=1pt},
        row{9} = {bg=orange!8, font=\itshape, rowsep=1pt},
    }
        \toprule
        \SetCell[r=2]{l, m} Method & \SetCell[c=4]{c} \textsc{MSR-VTT} & & & & \SetCell[c=4]{c} \textsc{PE-Video} & & & & \\
        \cmidrule[lr]{2-5}\cmidrule[lr]{6-10}
        & Rel. & Descr. & Temp C. & Flu.  & Rel. & Descr. & Temp C. & Flu.  \\
        \midrule
        \SetCell[c=9]{font=\itshape\bfseries} PerceptionLM & & & & & & & & & \\
        \quad Pretrained & 3.16 & 2.83 & 3.32 & 6.89 & 3.65 & 3.09 & 2.83 & 6.62 \\
        \quad SFT & 3.68 & 3.55 & 4.41 & 7.95 & 3.81 & 3.14 & 3.53 & 8.06 \\
        \quad DPO & 4.45 & 4.32 & 5.60 & 8.56 & 5.03 & 4.60 & 5.30 & 8.69 \\
        \quad MPO & 4.45 & 4.36 & 5.56 & 8.57 & 5.06 & 4.61 & 5.29 & \textbf{8.70} \\
        \quad Rank & \textbf{4.54} & \textbf{4.42} & \textbf{5.67} & \textbf{8.62} & \textbf{5.14} & \textbf{4.69} & \textbf{5.42} & 8.69 \\
        \midrule
        \SetCell[c=9]{font=\itshape\bfseries} Qwen2.5-VL & & & & & & & & & \\
        \quad Pretrained & 4.04 & 3.36 & 4.02 & 8.03 & 4.05 & 3.36 & 4.05 & 8.03 \\
        \quad SFT & 4.31 & 4.11 & 5.23 & \textbf{8.34} & 4.84 & 4.40 & 5.24 & 8.63 \\
        \quad DPO & 4.26 & 4.07 & 5.30 & \textbf{8.34} & 4.87 & 4.45 & 5.23 & \textbf{8.64} \\
        \quad MPO & 4.31 & 4.14 & \textbf{5.41} & 8.31 & 4.91 & 4.48 & 5.23 & 8.63 \\
        \quad Rank & \textbf{4.39} & \textbf{4.17} & \textbf{5.41} & \textbf{8.34} & \textbf{4.97} & \textbf{4.59} & \textbf{5.42} & 8.63 \\
        \bottomrule
    \end{tblr}
    \label{tab:captioning-gpt_oss}
\end{table*}

\subsection{Non-LLM-based Metrics}
While LLM-as-a-judge exhibits high correlations with human judgments and is a standard approach to perform captioning evaluations \cite{Matsuda2025vela,gu2024llm_as_judge_survey,Rawal2025ARGUSHA,Chai2024AuroraCapEP}, we acknowledge heavy usage of LLMs for both generating training data and evaluating the results. Therefore, outside of the human evaluation discussed in \cref{sec:experiments}, we provide the results for n-gram based metrics in \cref{tab:non_llmaj_results}. Higher values indicate greater overlap between the predicted and ground truth captions. \textbf{Notably, the ranking method achieves the best METEOR score and ties for the best ROUGE-L}, corroborating the LLM-based evaluations with traditional metrics. Finally, we want to highlight that MPO and DPO benefit from our dataset generated by our caption degradation method as they are fine-tuned on it.
\begin{table}[t]
    \small
    \centering
    \caption{Non-LLM-as-judge metrics for PLM on \cds-PVD. We note that DPO and MPO benefit from the data generated by our caption degradation method as they are fine-tuned on it. The Ranking provides further gains compared to MPO given that both are fine-tuned on the dataset generated by our method.}
    \begin{tblr}{
        colspec = {l * {3}{X[c]}},
        rowsep = .3pt,
        row{1} = {bg=gray!8}
    }
        \toprule
        \textbf{Method} & \textit{METEOR} & \textit{ROUGE-L} \\
        \midrule
        Pretrained & .152 & .234 \\
        SFT & .124 & .201 \\
        DPO & .196 & .292 \\
        MPO & .196 & \textbf{.293}\\
        Ranking & \textbf{.199} & \textbf{.293}\\
        \bottomrule
    \end{tblr}
    \label{tab:non_llmaj_results}
\end{table}

\section{Ranked Caption Chain Algorithm}
In \cref{fig:pseudocode}, we detail the algorithm to construct ranked caption chains for the \cds dataset. Since each error reliably lowers a caption's faithfulness to the video, the resulting captions form a clean, totally ordered chain.

\begin{figure}
    \begin{lstlisting}[
    language=Python,
    basicstyle=\ttfamily\footnotesize,
    showstringspaces=false,
]
def gen_caption_chain(
    orig_caption: str,
    video,
    chain_len: int
) -> list[str]:

    chain = [orig_caption]
    
    for i in range(chain_len):
        err_types = get_applicable_errors(chain[-1])
        err_type = sample(err_types)
        
        mutated_caption = apply_error_type(
            err_type,
            chain[:i],
            video
        )
        chain.append(mutated_caption)
    
    return chain

def apply_err_type(
    err_type, 
    chain: list[str], 
    video
) -> str:

    """
    Uses an LLM to apply an error type to the 
    latest caption in a chain while preserving 
    previously introduced errors.
    """
    ...
    \end{lstlisting}
    \caption{Algorithm to generate a chain of totally ordered captions from a starting caption. An LLM mutates the latest caption in the chain by introducing an error, preserving previously introduced errors and as much of the prior caption's structure as possible.}
    \label{fig:pseudocode}
\end{figure}

\section{Data Generation Design Choices}
In this section, we expand on some of the design choices touched upon in \textcolor{blue}{Table 1} in the main paper, providing additional details, motivations, and results.

\subsection{Effect of Error Sampling}
In the main paper, we noted that the PerceptionLM model trained on chains generated by our error-conditioned chain generation algorithm performed better than one trained on chains generated while allowing the LLM to freely insert errors. \cref{tab:free_err_gen} shows the results. The model trained on error-conditioned chains \textbf{performs better across all four metrics}, suggesting that diversity in introduced errors helps the model obtain broad performance gains.

\begin{table}[]
    \centering
        \caption{Comparison between PLM trained on chains generated with error-conditioning vs.\ PLM trained on chains with freely introduced errors. 
        }
        \begin{tblr}{
            colspec=lccccc,
            rowsep=1pt,
            row{1-2}={bg=gray!9},
            row{2}={font=\itshape},
            hline{2}={3-6}{leftpos=-1,rightpos=-1, endpos},
        }
            \toprule
            \SetCell[r=2]{} \textbf{Model} & \SetCell[r=2]{} {\bfseries Error \\Condition.} & \SetCell[c=4]{} \cds-PVD & & & \\
            & & Relevance & Descriptiveness & Temporal C. & Flu. \\
            \midrule
            PerceptionLM & \ding{55} &  7.48 & 7.41 &  7.62 &  8.80 \\
            PerceptionLM & $\checkmark$ &  \textbf{7.67} & \textbf{7.72} & \textbf{7.93} & \textbf{9.49} \\
            \bottomrule
        \end{tblr}

    \label{tab:free_err_gen}
\end{table}

\subsection{Effect of Conditional Caption Generation}
A key element of our data generation method is that each generated caption is the same as the one prior, except with an additional error. So the caption $c_k$ at rank $k \ge 1$ has $k - 1$ errors in it, and the set of errors of caption $k$ is a strict subset of those of the next: $\operatorname{Errors}(c_k) \subsetneq \operatorname{Errors}(c_{k + 1})$. 

One might wonder whether maintaining the set of previous errors provides any benefit. For example, instead of caption $c_{k+1}$ containing all of the errors of $c_k$, we could instead have each caption contain a single error introduced into the original caption, independent of all other captions. Thus, we would have $c_1$ as the original caption and $c_k$ for $k \ge 2$ each with a single error independent of the other captions. One might expect, also, that training on these negatives would yield better performance due to their each being closer to the original caption than progressively degrading captions, thereby serving as harder negatives.

To evaluate this setting, we generate these ``independent negatives'' using GPT-OSS 20B and train PLM with MPO to ensure all captions in the set are used. We compare the model to PLM trained with ranking on the caption chains in \cref{tab:indep_errors} with GPT-OSS as a judge. This baseline is described as ``error injection'' in \cref{tab:data_gen} (results repeated here for completeness).

\begin{table}[]
    \centering
    \caption{Comparison between progressively degraded caption chains vs.\ sets of captions where each caption contains a single, independent error based on the original caption. Captions are generated and evaluations performed by GPT-OSS 20B. \textbf{Progressive degradation with ranking outperforms independent errors across all metrics.}}
    \begin{tblr}{
        colspec = lcccc, 
        row{1} = {bg=gray!9}
    }
        \toprule
        \textbf{Model} & \textit{Relevance} & \textit{Descriptiveness} & \textit{Temporal Consist.} & \textit{Fluency} \\
        \midrule
        MPO (Indep.\ Errs) &  4.90 & 4.62 & 5.38 & 8.46 \\
        Ranking & \textbf{5.27} & \textbf{4.92} & \textbf{5.75} & \textbf{8.47} \\
        \bottomrule
    \end{tblr}
    \label{tab:indep_errors}
\end{table}

\subsection{On-Policy Evaluation}
One major element of our proposed method is that data is generated off-policy. While off-policy data's reusability across models makes it highly efficient, prior work has shown that on-policy data drives strong performance which can outperform off-policy DPO \cite{yang2025onpolicydpo}.

To compare our proposed off-policy ranking framework with on-policy methods, we train several baselines with on-policy data generated using our PE-Video training set. \cite{yang2025onpolicydpo} describes two classes of on-policy DPO methods. The first uses an external LLM judge to improve existing generations, with the improved generations serving as the preferred responses in DPO. The second class uses an external LLM judge to perform pairwise comparisons of the target model's responses, constructing preference pairs from the existing responses. For this evaluation, we include representative methods from each of these classes, each of which is described below. Each method starts by constructing initial on-policy captions, generating five captioning responses per video. 
\paragraph{Response Improvement.} We take each of the five generated responses and have GPT-OSS 20B \cite{IntroducingGPTOSS2025} refine each of the responses based on the ground truth caption, fixing errors and incorporating missed elements. This process yields five win/loss pairs which we use for our DPO training data.
\paragraph{Response Comparison.} We take all ten combinations of caption pairs and feed them to GPT-OSS 20B, asking it to indicate which caption better represents the video given the ground truth caption. GPT-OSS is allowed to say that the captions are of approximately equal quality, in which case a preference pair is not generated. The vast majority of videos generate nine or ten caption pairs, so we select all examples with at least nine preference pairs, sampling nine such pairs per video (using a uniform number of pairs allows for efficient distributed training).
\paragraph{RLAIF-V.} RLAIF-V \cite{yu2025rlaifv} uses an LLM to extract claims from each response, then evaluates each of the claims' correctness to produce a score for each response based on the number of incorrect claims. Preference pairs are then constructed between captions with differing scores. This method is similar to our proposed approach, which orders captions based on the number of introduced errors. 

As this method is only designed to reduce hallucinations by evaluating responses based on their precision, we extend the RLAIF-V framework to also compute the number of ground truth claims entailed by the response (recall). We then compute each response's score as the harmonic mean of these values (F1 score). We call this version \textit{Bidirectional} RLAIF-V (Bidir).

\subsubsection{Evaluation.} We use GPT-OSS 20B to recaption the PE-Video dataset and regenerate chains for our ranking method. All models are trained for 500 optimization steps with a batch size of eight videos using DPO/ranking loss and next-token prediction on the ground truth caption with a weight of .1. For each of the on-policy methods, we also experiment with performing a pre-SFT step on the ground truth captions for on-policy adaptation prior to generating initial responses \cite{yang2025onpolicydpo}.

\cref{tab:on_policy} shows the results on PE-Video captioning, with evaluations performed by GPT-OSS 20B as a judge. While all methods significantly improve performance over the baseline, \textbf{our off-policy ranking approach demonstrates the strongest performance on relevance, descriptiveness, and temporal consistency}. We attribute this both to the challenge that the ranking task poses, and to the next-token prediction loss, which serves both to prevent generation collapse and to shift the preferred response to be in-distribution \cite{yang2025onpolicydpo}. LLM-based response refinement is the best-performing on-policy baseline, followed by response comparison and RLAIF-V. RLAIF-V has much lower \textit{descriptiveness} as it is a primarily hallucination-centric method---its response scores are based on the number of incorrect claims made in a response. Our bidirectional variant improves its scores on three out of four metrics, but it still does not outperform ranking on those same metrics.

\begin{table*}[h!]
    \small
    \centering
    \caption{Comparison between on- and off-policy data generation methods training PLM on \cds-PVD. Despite having model-tailored data generation, the on-policy baselines do not outperform our ranking approach on video detailed captioning. Data generated and evaluated with GPT-OSS 20B.}
    \begin{tblr}{
        colspec = {X[3,l] *{5}{X[1,c]}},
        colsep = 4pt,
        rowsep = 1pt,
        row{1} = {font=\bfseries, bg=gray!8},
        row{2} = {font=\itshape, bg=gray!8},
        row{3} = {bg=blue!8, font=\itshape, rowsep=1pt},
    }
        \toprule
        \SetCell[r=2]{l, m} Method & \SetCell[r=2]{c,m} On-Policy & \SetCell[c=4]{c} \textsc{PE-Video} & & & \\
        \cmidrule[lr]{3-6}
        & & Rel. & Descr. & Temp C. & Flu.  \\
        \midrule
        \SetCell[c=5]{font=\itshape\bfseries} PerceptionLM & & & & & \\
        Pretrained & \ding{55} & 3.65 & 3.09 & 2.83 & 6.62 \\
        Rank & \ding{55} & \textbf{5.27} & \textbf{4.92} & \textbf{5.75} & 8.47 \\
        \hline
        Response Comparison & $\checkmark$ & 5.03 & 4.79 & 5.51 & 8.45 \\
        \quad With Pre-SFT & $\checkmark$ & 5.12 & 4.71 & 5.47 & 8.60 \\
        Response Improvement & $\checkmark$ & 5.07 & 4.83 & 5.57 & 8.52 \\
        \quad With Pre-SFT & $\checkmark$ & 5.07 & 4.92 & 5.32 & 7.61 \\
        RLAIF-V & $\checkmark$ & 5.06 & 4.35 & 5.21 & \textbf{8.67} \\
        \quad With Pre-SFT/Bidir & $\checkmark$ & 5.13 & 4.83 & 5.52 & 8.57 \\
        \bottomrule
    \end{tblr}
    \label{tab:on_policy}
\end{table*}

\section{Validity of Chain Order}
Generated chains are totally ordered by construction, as each successive caption has the same content as the one prior (maintaining existing errors), but with the addition of one extra error. However, order can be disrupted if the LLM fails to maintain the prior text or inadvertently makes a caption more faithful. To quantify chain validity, we sample 100 chains from \cds-PVD for human inspection, finding that \textbf{89\% of chains maintain the total order}. Moreover, 8 out of 11 failures are due to errors already present in the original caption. This means that \textbf{only 3\% of the chains contain noise introduced by our data generation approach}, highlighting our method's robustness.

\section{Choice of Ranking Loss}
In the main text, we compared the Plackett-Luce loss to Hinge loss due to the latter's not enforcing a margin. Here, we perform an additional experiment using the RankNet loss of \cite{burges2005ranknet}. Models are trained on GPT-OSS chains and evaluated by the same model. We find that RankNet outperforms Hinge loss but performs worse than Plackett-Luce. As the RankNet loss is essentially DPO without a reference model, this suggests that \textbf{the KL constraint is a beneficial element of the Plackett-Luce formulation}.

\begin{table}[h!]
    \small
    \centering
    \caption{Comparison of ranking losses with PLM on \cds-PVD. The Plackett-Luce loss achieves the best results on three out of four metrics.}
    \begin{tblr}{
        colspec = {l *{4}{X[c]}},
        colsep = 4pt,
        rowsep = .3pt,
        row{1} = {font=\itshape, bg=gray!8},
    }
        \toprule
        \SetCell{font=\bfseries} Method& Relevance & Descriptiveness & Temporal C. & Fluency  \\
        \midrule
         RankNet~\cite{burges2005ranknet} & 4.91 & 4.67 & 5.42 & \textbf{8.52} \\
         Hinge & 4.86 & 4.63 & 5.36 & 8.43 \\
         Plackett-Luce & \textbf{5.27} & \textbf{4.92} & \textbf{5.75} & 8.47 \\
        \bottomrule
    \end{tblr}
    \label{tab:ranking_loss}
\end{table}

\section{Captioning Example}
\begin{figure*}
    \centering
    \includegraphics[width=.9\textwidth]{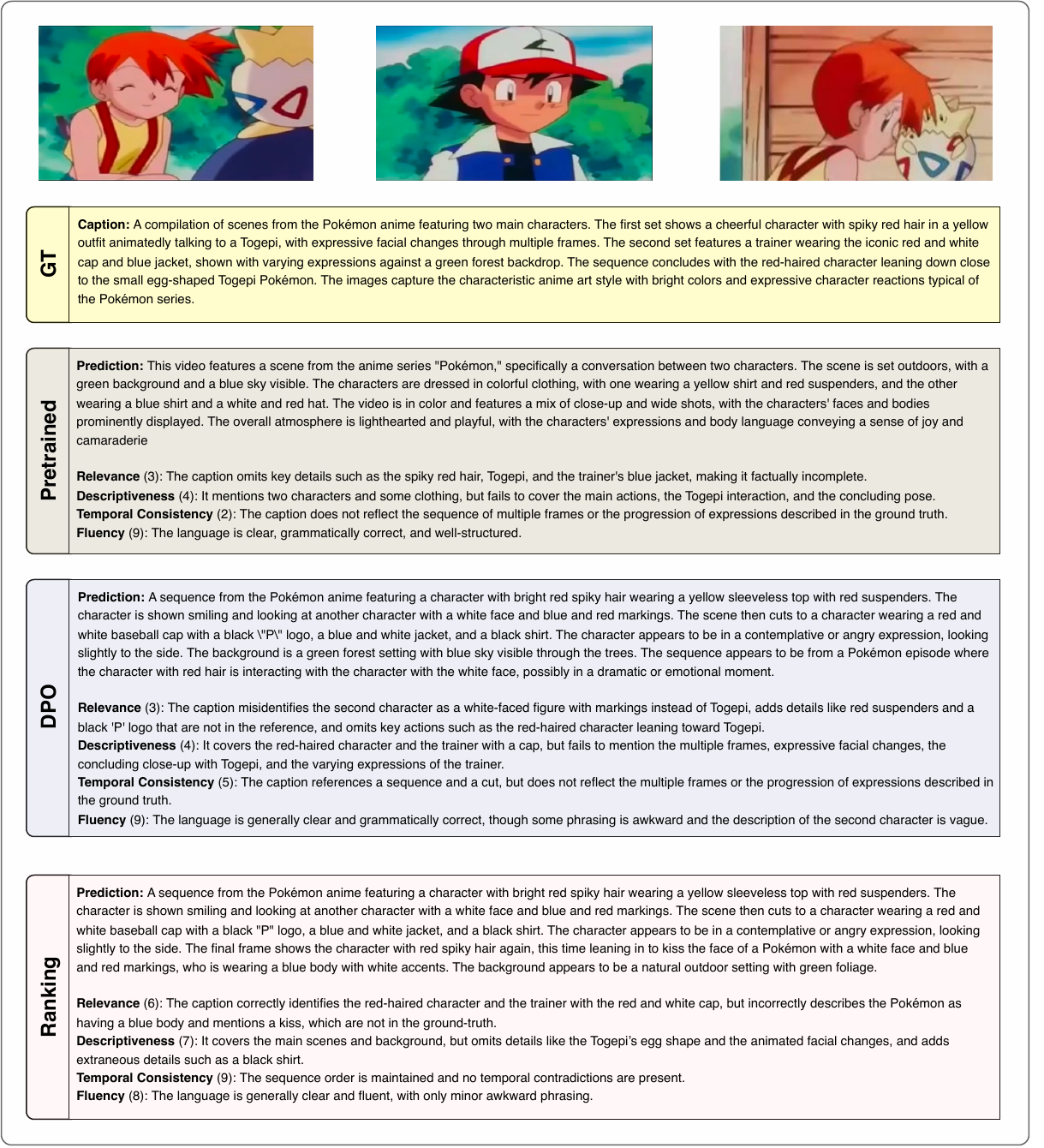}
    \caption{An example of model caption predictions on MSR-VTT with GPT-OSS 20B as a judge.}
    \label{fig:cap_example1}
\end{figure*}

\cref{fig:cap_example1} provides an example of model-generated captions on an MSR-VTT video with GPT-OSS 20B as a judge. While the DPO and Ranking models are more informative than the pretrained model's description, both contain inaccuracies highlighted by the LLM judge.

\section{Examples of Long-Form Multiple Choice Questions}
\cref{fig:mcq_examples} provides examples from our Long-Form Multiple Choice Question Answering benchmark. Note the questions' complexity and the long-form nature of each of the answer choices, the latter of which challenges models' attention to visual detail.

\begin{figure*}[h]
    \centering

    \begin{subfigure}{0.95\textwidth}
        \centering
        \includegraphics[width=\textwidth,height=0.9\textheight,keepaspectratio]{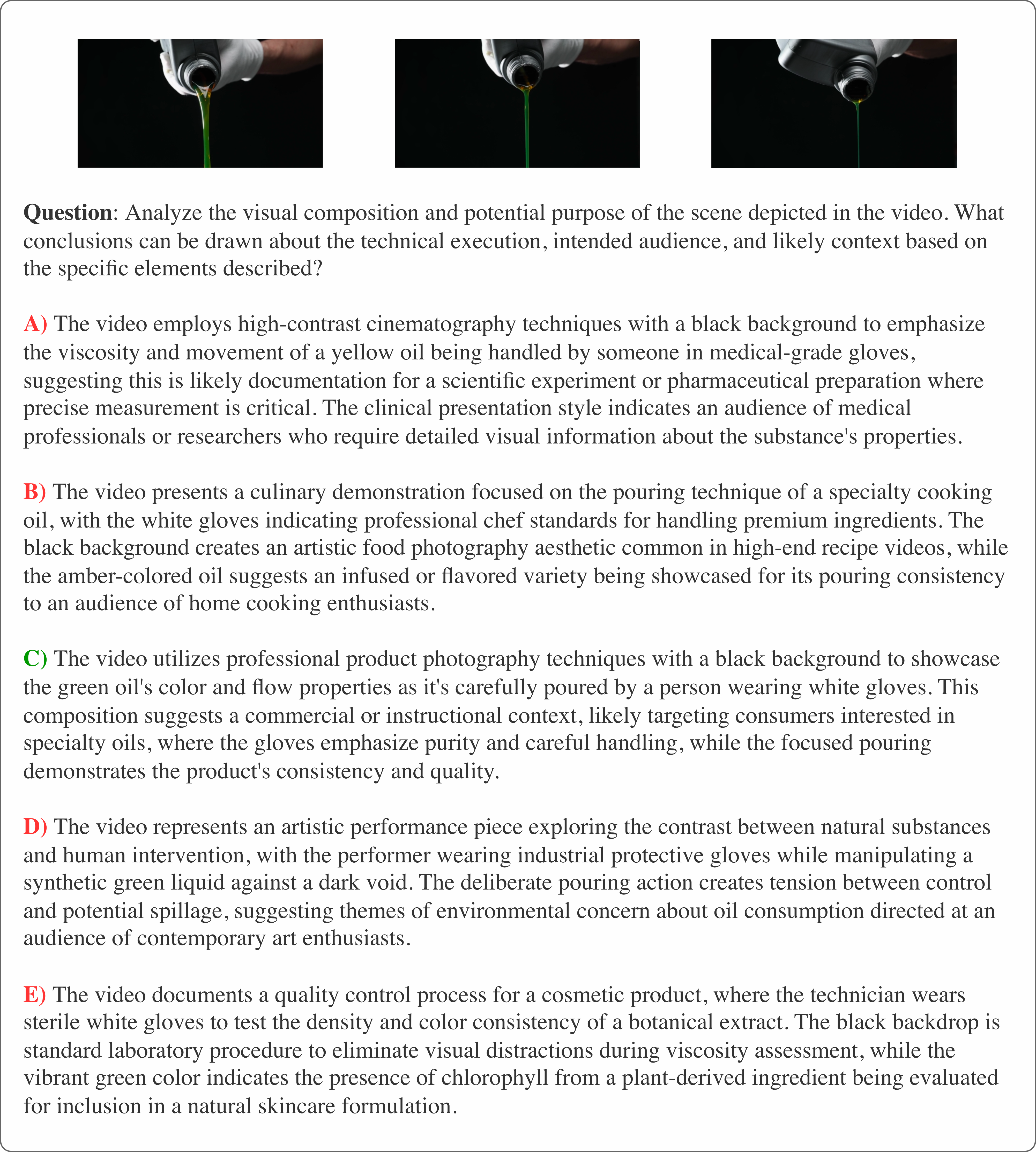}
        \label{fig:mcq_example_1}
    \end{subfigure}

    \caption{Examples from our Long-Form Multiple Choice Question Answering benchmark. Green represents the correct answer.}
    \label{fig:mcq_examples}
\end{figure*}

\begin{figure*}[h]
    \ContinuedFloat
    \centering

    \begin{subfigure}{0.95\textwidth}
        \centering
        \includegraphics[width=\textwidth,height=0.9\textheight,keepaspectratio]{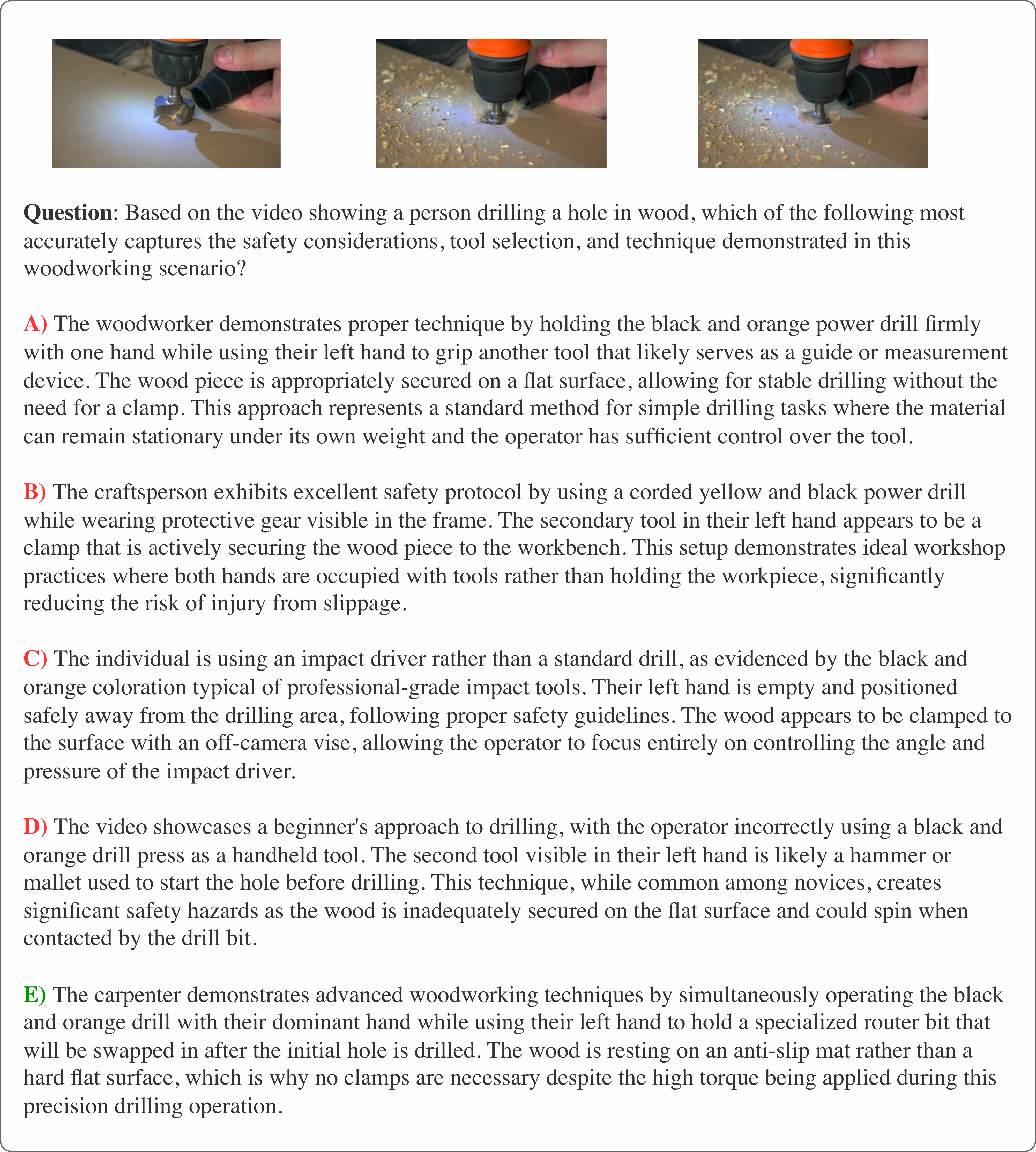}
        \label{fig:mcq_example_2}
    \end{subfigure}
\end{figure*}

\ansel{Potentially include more implementation details? Including prompts for data generation}

\section{Limitations}
While our proposed method demonstrates scalability and strong performance over close alternatives, it is not without drawbacks.

First, our method generally assumes access to detailed captions to ensure we can generate a chain with an increasing number of errors. While this process results in cleanly ordered captions, it may require recaptioning of existing datasets so each caption contains sufficient claims that can be modified. Alternatively, one may repeatedly modify a single statement to become less and less faithful to the visual inputs, as demonstrated in \cref{sec:transforming_captions}. This tends to generate less cleanly ordered chains, however, as introducing errors into a long caption is a much easier task for an LLM than is repeatedly making an existing statement less faithful to the video.

One key property of our method is that the caption chain data is generated off-policy. This has both advantages and drawbacks. By generating model-independent caption chains, one is able to rapidly fine-tune new models on the data without needing to regenerate model-specific data. \cref{sec:experiments} shows that these chains are useful across two strong model families, despite the data being generated off-policy. Further, the fact that our data generation framework is off-policy allows one to bias the generated data to target particular weak points by oversampling certain error types. On the other hand, on-policy DPO has generally been shown to be more effective than off-policy DPO, at the cost of regenerating data on a per-model basis \cite{yang2025onpolicydpo,shenfeld2025rlsrazor}. Adapting our method to modify model-generated captions in an on-policy manner constitutes a promising direction for future research.

\bibliographystyle{splncs04}
\bibliography{references}

%% file: sec/preamble.tex
\usepackage{commath}
\usepackage{xparse}
\usepackage[dvipsnames]{xcolor}
\usepackage{multirow}
\usepackage{tabularray}
\UseTblrLibrary{booktabs}
\usepackage{xspace}
\usepackage{mathtools}
\usepackage{listings}
\usepackage{float}
\usepackage{pifont} 
\usepackage{enumitem} 

\makeatletter
\let\tblr@booktabs@hline\tblr@booktabs@oldhline
\let\tblr@booktabs@cline\tblr@booktabs@oldcline
\let\tblr@booktabs@cline@more\tblr@booktabs@oldcline@more
\makeatother

\NewDocumentCommand{\ansel}{m}{}

\newcommand{\cds}{\textbf{\textsc{Rcc}}\xspace}

\usepackage{tikz}
\usepackage{xcolor}
\usepackage{fontawesome5}
\usetikzlibrary{positioning, calc, backgrounds}

\definecolor{attrC}{HTML}{E87040}
\definecolor{objC} {HTML}{3B52A5}
\definecolor{actC} {HTML}{3DAA6A}
\definecolor{spatC}{HTML}{8B55B8}
\definecolor{refC} {HTML}{D44F6E}
\definecolor{tempC}{HTML}{3FADA6}

\definecolor{attrBg}{HTML}{FEF5EC}
\definecolor{objBg} {HTML}{EEF1FA}
\definecolor{actBg} {HTML}{EEF8F2}
\definecolor{spatBg}{HTML}{F5EEF9}
\definecolor{refBg} {HTML}{FAEEF2}
\definecolor{tempBg}{HTML}{EEF9F7}

\definecolor{outerBg}{HTML}{E4E9EF}
\definecolor{sepCol} {HTML}{BBBBBB}

%% file: sec/abstract.tex
\begin{abstract}
Direct preference optimization (DPO) is an effective technique to train language models to generate preferred over dispreferred responses. However, this binary ``winner-takes-all'' approach is suboptimal for vision-language models whose response quality is highly dependent on visual content. In particular, a response may still be faithful to the visual inputs even if it is less preferable than an alternative. The standard Bradley-Terry DPO formulation lacks this nuance, upweighting winning responses without sufficient regard for whether the ``losing'' response still maintains high visual fidelity. In this work, we investigate ranking optimization as an alternative that more precisely situates responses' faithfulness to visual inputs. We focus on video-text alignment using detailed video captions, proposing a method to generate challenging, totally ordered caption chains at scale through repeated caption degradation.  Our results show ranking optimization outperforms binary DPO for long-form content generation and assessment, and importantly, we find that these approaches require finetuning of the vision encoder to be effective, challenging the view of DPO as purely a language-reweighting process.

\end{abstract}

%% file: sec/intro.tex
\section{Introduction}

\begin{figure*}[t]
\centering
\begin{tikzpicture}[
    font=\sffamily\footnotesize,
    caption/.style={
        rectangle,
        draw=black!50,
        fill=white,
        rounded corners=2pt,
        text width=2.0cm,
        align=left,
        drop shadow,
        inner sep=1.5pt,
        line width=0.5pt,
        font=\sffamily\tiny
    },
    arrow/.style={
        -{Stealth[length=1.5mm, width=1mm]},
        line width=1pt,
        draw=black!70
    },
    label/.style={
        rectangle,
        rounded corners=2pt,
        fill=blue!10,
        draw=blue!50,
        font=\bfseries\footnotesize,
        inner sep=3pt,
        drop shadow
    },
    vlmbox/.style={
        rectangle,
        rounded corners=4pt,
        fill=blue!5,
        draw=blue!70,
        line width=1.5pt,
        inner sep=4pt,
        drop shadow,
        minimum width=1.5cm,
        minimum height=0.8cm
    },
    scale=0.8,
    transform shape
]

\node[inner sep=0pt, drop shadow] (img) at (-5.2, 0) {
    \includegraphics[width=2.2cm]{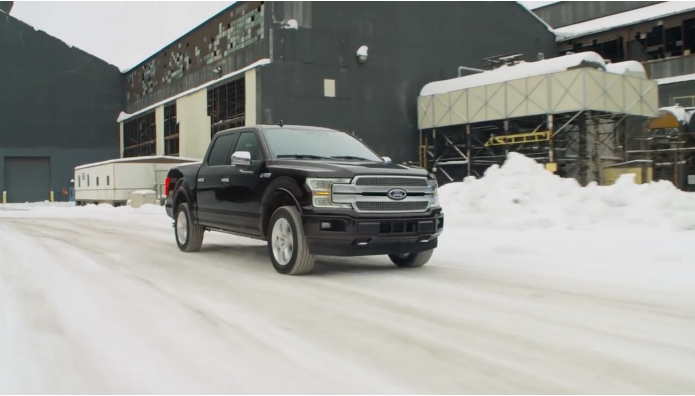}
};

\node[circle, fill=black!70, inner sep=0pt, minimum size=0.5cm, opacity=0.7] at (img.center) (playbutton) {};
\node[regular polygon, regular polygon sides=3, fill=white, inner sep=0pt, minimum size=0.2cm, 
      shape border rotate=90] at (playbutton.center) {};

\node[caption] (cap1) at (-2.2, 1.5) {
    \textbf{Cap 1:} A powerful pickup truck navigates through a snow-covered industrial 
    area on a \textbf{winter} day. .....
    handle \textbf{harsh} winter environments.
};

\node[caption] (cap2) at (-2.2, 0.0) {
    \textbf{Cap 2:} A powerful pickup truck navigates through a snow-covered industrial 
    area on a \textbf{\color{red}sunny} day. .....
    handle \textbf{harsh} winter environments.
};

\node[caption] (cap3) at (-2.2, -1.60) {
    \textbf{Cap 3:} A powerful pickup truck navigates through a snow-covered industrial 
    area on a \textbf{\color{red}sunny} day. ....
    handle \textbf{\color{red}easy} winter environments.
};

\node[right=0.05cm of cap1.east, anchor=west, font=\sffamily\tiny\itshape, text=black!60] (elabel1) {GT};
\node[right=0.05cm of cap2.east, anchor=west, font=\sffamily\tiny\itshape, text=black!60] (elabel2) {1 Err};
\node[right=0.05cm of cap3.east, anchor=west, font=\sffamily\tiny\itshape, text=black!60] (elabel3) {2 Err};

\node[label, font=\bfseries\small] (query) at (1.5, 1.8) {\textbf{Query:} Rank captions};

\node[vlmbox] (vlmbox) at (1.5, 0) {
    \textbf{\large VLM}
};

\node[label, fill=red!10, draw=red!50, font=\bfseries\scriptsize] (rank1) at (4.2, 0.6) {
    \begin{tabular}{c}
    \textbf{Prediction:} \\
        \textcolor{red!70}{1, 3, 2} \textcolor{red!70}{\bfseries \ding{55}}
    \end{tabular}
};

\node[label, fill=green!10, draw=green!50, font=\bfseries\scriptsize] (rank2) at (4.2, -0.6) {
    \begin{tabular}{c}
         \textbf{Ground Truth:} \\
        \textcolor{green!50!black}{1, 2, 3} \textcolor{green!50!black}{\bfseries \ding{51}}
    \end{tabular}
};

\draw[arrow] (img.south) -- ++(0, -1.84) -| (vlmbox.south);

\draw[arrow] (elabel1.east) to[out=0, in=160] (vlmbox.north west);
\draw[arrow] (elabel2.east) to[out=0, in=180] (vlmbox.west);
\draw[arrow] (elabel3.east) to[out=0, in=200] (vlmbox.south west);

\draw[arrow] (query.south) -- (vlmbox.north);

\draw[arrow, red!70] (vlmbox.east) to[out=0, in=180] (rank1.west);

\end{tikzpicture}
\caption{Existing models struggle to differentiate between visual details in long-form and detailed captions. We propose caption ranking to capture fine-grained differences for video-language alignment.}
\label{fig:teaser}
\vspace{-1em}
\end{figure*}

Vision-language models (VLMs) have made great strides in recent years, driving progress in embodied reasoning and a wide range of visual understanding tasks, including retrieval, detection, segmentation, and video analysis.\cite{Kim2024OpenVLAAO,Driess2023PaLMEAE,Chen2023InternVS,Lin2023VideoLLaVALU,Rasheed2023GLaMMPG}. Despite their strengths, VLMs often struggle to understand fine-level visual details, spatial relationships, and object parts, and may have difficulty identifying straightforward actions in videos \cite{Kim2024Finer,Tong2024eyeswideshut,Blume2025PARTONOMYLM,Wang2023PaxionPA}. This is in part because they are insensitive to details in their visual and textual inputs, causing them to miss fine-level information and generate inaccurate statements \cite{Kim2024Finer,Yuksekgonul2022when_and_why,Liu2023Lostinthemiddle}.

Techniques to improve VLMs' vision-language alignment diverge depending on the VLMs' architectures. On the one hand, CLIP-style \cite{Radford2021clip}, dual-encoder architectures are contrastively trained using in-batch negatives, with some works including specially selected hard negatives to develop robust, language-aligned visual representations~\cite{radenovic2023filtering,xie2025fg}. On the other hand, generative VLMs typically borrow post-training concepts like PPO \cite{Schulman2017ppo}, DPO \cite{Rafailov2023dpo}, and GRPO \cite{Shao2024grpo} from the LLM literature. Existing works on generative VLM alignment typically apply these techniques by freezing the vision encoder and tuning the LLM, approaching issues like hallucination from a purely language modeling perspective.

Current approaches to VLM post-training suffer from a limitation in their ability to distinguish between fine-grained visual and textual details. For example, assume for a video we have a ground truth caption \emph{a dog playing with a toy}, and two negative captions (1) \emph{a cat playing with a toy} and (2) \emph{a cat playing with a child}. Both captions contain inaccurate information, but caption (1) is more accurate than caption (2) in light of the reference caption. The binary training objective implemented most often for DPO, where a caption is either positive or negative, cannot capture nuances in caption quality, instead treating both minor misrepresentations and fully unrelated captions as ``incorrect''.

Motivated by the limits of this training paradigm, we propose a caption-ranking approach that trains models to attend to increasingly fine-grained visual details. To achieve this, we focus on video detailed captioning \cite{Chai2024AuroraCapEP}, where a model must generate a detailed caption for a provided video. Indeed, a model which is able to produce an accurate and comprehensive caption has a strong understanding of its visual inputs. Our approach leverages an LLM to create ranked caption sequences, beginning with rich, comprehensive descriptions that are systematically mutated to produce lower-quality variants. This iterative process generates captions with high similarity to force models to attend to differences in visually-grounded details.
\begin{figure*}[h]
    \vspace{-1em}
    \centering
    \includegraphics[width=.90\linewidth]{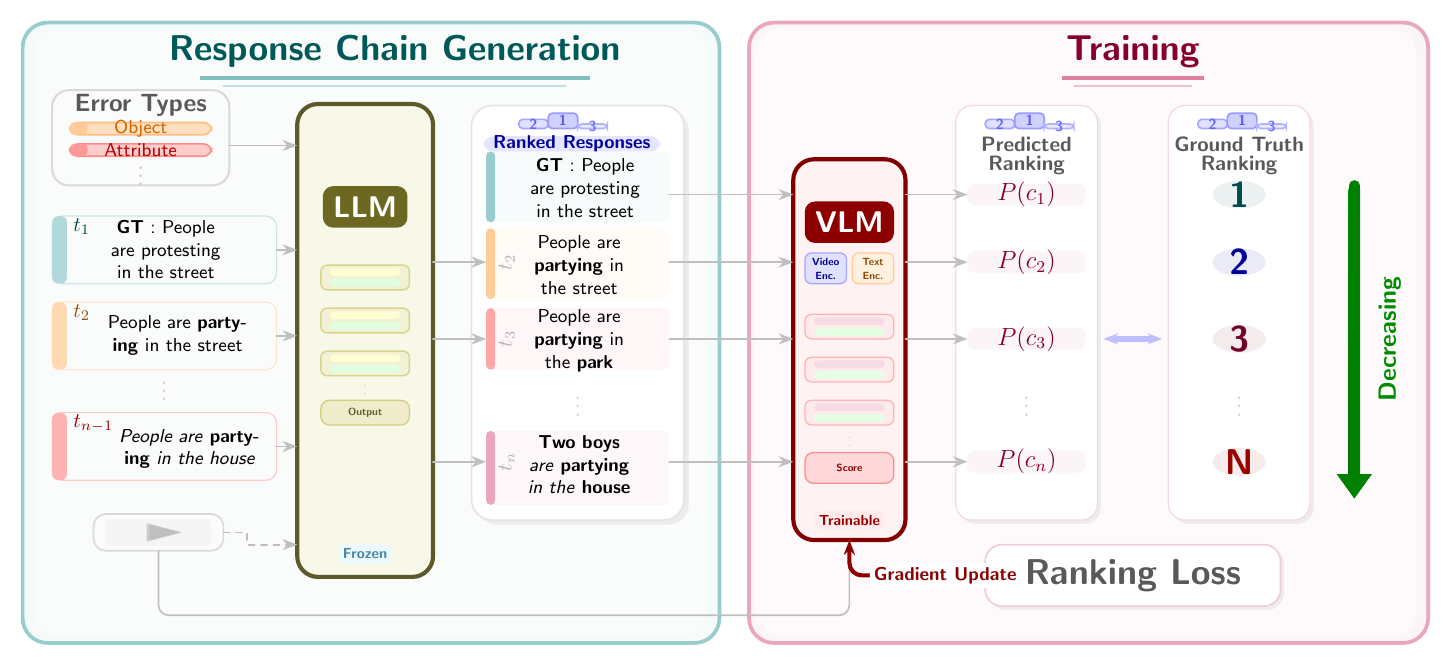}
    \caption{Our proposed framework. Given a set of high-quality video captions, an LLM generates a totally ordered caption chain by repeatedly introducing visually-grounded errors into the previous caption. These caption chains are then used to train VLMs by having them rank the generated captions using ranking-based DPO. The captions' similarity forces the models to pay attention to fine, visually-grounded details.}
    \label{fig:workflow}
    \vspace{-1em}
\end{figure*}
We argue that an ideal objective should reflect the nuanced reality that \textit{some captions are better than others} rather than being simply correct or incorrect. To achieve this, we employ the Plackett-Luce ranking formulation of DPO \cite{Rafailov2023dpo} and train models with our generated caption chains.

We benchmark our generated chains against standard on- and off-policy approaches and evaluate ranking against three baselines: binary DPO, supervised fine-tuning (SFT), and Multi-Preference Optimization (MPO), which uses multiple negatives. Our evaluation spans three types of benchmarks: (1) detailed video caption generation tasks (MSRVTT, PVD, VDC, and ARGUS) \cite{Xu2016MSRVTTAL,Bolya2025PerceptionET, Rawal2025ARGUSHA, Chai2024AuroraCapEP}, (2) a custom Long-Form Multiple Choice Question Answering benchmark designed to assess fine-grained comprehension of question and answer choices, and (3) a caption matching task (TempCompass) \cite{liu2024tempcompass}.

Our results show that standard Bradley-Terry DPO improves captioning quality over SFT, and that ranking outperforms both DPO and MPO. We further observe that these gains from ranking may only be achieved if the \emph{vision encoder is finetuned with the LLM backbone}, suggesting that the widespread language-focused perspective of DPO may not realize its full potential. In summary, our contributions are as follows:
\begin{enumerate}
    \item We identify a deficiency in the use of DPO for vision-language alignment and propose to use a ranking objective to ameliorate it.
    \item We propose a degradation paradigm for generating ranked caption chains at scale, enabling effective ranking optimization. We contribute the \cds Ranked Caption Chain dataset generated using this approach.
    \item We demonstrate how to transform caption chains into multiple choice and yes/no questions, extending the ranking paradigm to question answering.
    \item We perform comprehensive experiments demonstrating ranking's effectiveness for vision-language alignment, we thoroughly explore the design space of our ranking framework, and we provide a new perspective on using DPO for visual representation learning.
\end{enumerate}

%% file: sec/related_work.tex
\section{Related Work}
\paragraph{DPO for Vision-Language Models.}
Direct Preference Optimization (DPO) has emerged as a prominent technique for mitigating hallucination in Vision-Language Models (VLMs)~\cite{ouali2024clip,Ding2025PaMiVDPOMV,xie2024v,zhang2025direct,poppi2026countervid}. CLIP-DPO~\cite{ouali2024clip} employs GPT-4 to systematically inject hallucinations into image-caption pairs, targeting three distinct categories: object existence, attribute misrepresentation, and relational inaccuracies. The method subsequently applies a standard DPO objective for model refinement. By contrast, V-DPO~\cite{xie2024v} adopts a visual manipulation approach, utilizing Stable Diffusion to modify objects referenced in captions, thereby creating preference pairs for DPO-based training. MDPO leverages model-generated captions alongside GPT-augmented outputs to construct preference pairs, enabling self-supervised refinement through the DPO framework.

\paragraph{Contrastive Learning Paradigms.}
Contrastive learning has been used predominantly in self-supervised representation learning for vision encoders~\cite{he2020momentum,chen2020improved,chen2020simple}. The methodology relies on constructing positive pairs through systematic data augmentation while deriving negative pairs from distinct samples within mini-batches or memory banks. CLIP and related models \cite{Radford2021clip,zhai2023siglip,jia2021align,Huang2024llm2clip} perform vision-language alignment through contrastive learning, aligning an image with its caption and treating other in-batch images and captions as negatives. DPO can be analogized as contrastive learning for generative language models, increasing the likelihood that one response is generated over another.

\paragraph{Video Captioning Methods.}
Distinct from existing DPO applications targeting image-text alignment, our approach addresses the more complex challenge of video-text alignment. Contemporary video captioning research has primarily concentrated on architectural innovations~\cite{zhou2024streaming,islam2024video,xu2024retrieval,kim2024you} or leveraging proprietary models for enhanced caption generation ~\cite{chen2024sharegpt4video,chen2024panda,zhao2024distilling}. By contrast, we contribute a novel framework that generates ranked caption chains through LLM-based degradation, advancing fine-grained, long-form text comprehension and generation capabilities in video-language models.

%% file: sec/method.tex
\section{Learning to Rank Captions}

\subsection{Data Generation}
The primary challenge in training models to rank is obtaining high-quality ranking data. Prior work \cite{Ouyang2022instructGPT} obtains ranking data by having human annotators rank model outputs for a fixed prompt. They then train a reward model on all $\binom{k}{2}$ pairwise preferences. This approach has two key drawbacks. First, it is time-consuming and expensive to have human annotators generate ranking data at scale. An automated framework that produces high-quality rankings would be preferable. Second, model responses may admit only a partial order. Consider the following hypothetical generations for an image of an empty table: ``A bowl of fruit on a table'', ``A pitcher of water on a table''. Both answer choices have errors, and it is unclear which is better. Enforcing an ordering upon potentially incomparable responses introduces noise into the training process.

With these challenges in mind, we propose to generate ranking data by using an LLM to progressively degrade captions. We begin the chain generation process with a ground truth caption $c_1$. To generate the caption $c_{k + 1}$ from $c_k$, we start by sampling an error type from a pre-defined set of visually-grounded errors. An LLM then mutates $c_k$ to introduce an error of the selected type, while ensuring that the mutated caption $c_{k+1}$ maintains $c_k$'s structure and all previously introduced errors. This mutation procedure is repeated to generate a chain of captions of decreasing quality, so that $c_1 \succ c_2 \succ \ldots \succ c_n$. This generation process has the following advantages:
\begin{enumerate}[leftmargin=2\parindent]
    \item \textbf{Generated chains are totally ordered}. A caption's rank corresponds to the number of errors it contains, and caption $c_{k+1}$ includes all of $c_k$'s errors. Hence, the ordering is well-defined and generated chains are totally ordered by their faithfulness to the visual inputs.
    
    \item \textbf{Captions in a chain have consistent structure}. As the mutation process maintains previous captions' structure, we ensure that the relative likelihood of one caption to another is solely due to the captions' introduced errors, and not due to linguistic variation (e.g. differences in phrasing, length).
   
    \item \textbf{Control over the error distribution}. Sampling from a pre-defined set of error types ensures one may diversify the introduced errors or focus sample generation on areas a model struggles on. Early experiments found that allowing the LLM to freely introduce errors biased the distribution towards object and attribute-centric alterations.
    
    \item \textbf{Chains can be safely generated at scale.} Due to the simplicity of the mutation step---one needs only modify the caption to introduce an error---chain generation can be performed at scale by an LLM.
\end{enumerate}
As a caption's rank in a chain corresponds to the number of errors it contains, we require that the ground truth caption $c_1$ be sufficiently descriptive. This allows one to introduce errors by modifying the caption instead of extending it, maintaining length consistency across the chain.

\subsection{The Ranked Caption Chain (RCC) Dataset}
\label{sec:rcc}
\begin{figure*}[!b]
    \centering
    \includegraphics[width=0.7\linewidth]{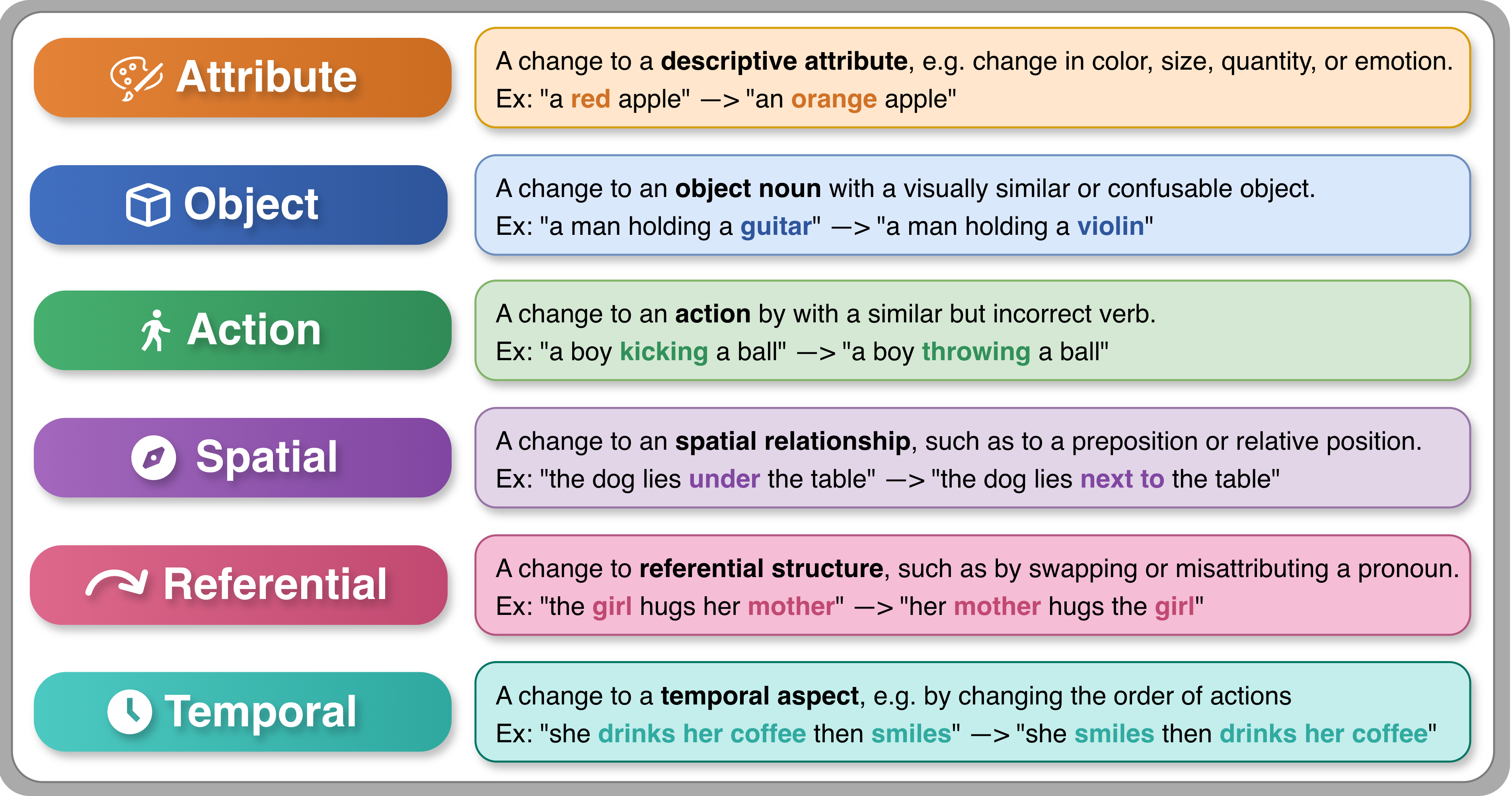}
    \caption{Error types used to generate caption chains through repeated caption mutation.}
    \label{fig:error_types}
\end{figure*}

We apply our data generation procedure to subsets of the PE-Video (PVD) \cite{Bolya2025PerceptionET} and MSR-VTT \cite{Xu2016MSRVTTAL} datasets, creating the \cds-PVD and \cds-MSR datasets. 

Before chain generation, we ensure that the starting captions for each chain are of sufficient length by having the frontier LLM Claude 3.7-Sonnet recaption them. For a given example, we provide Claude with the human-annotated captions (PE-Video has one per video, MSR-VTT has multiple), prompting it to use the provided captions and video to generate a detailed video caption. Manual inspection shows that providing the ground truth captions mitigates hallucination, resulting in high-quality detailed captions. 

To maximize the controllability of our chain generation process, we define a set of error types that we have Claude apply to captions (\cref{fig:error_types}). These errors are designed to capture a wide range of visually-grounded inaccuracies that one may find in video captions. We use part-of-speech tagging and keyword-based heuristics to determine the set of error types that are applicable to a caption. We sample from this filtered set uniformly before prompting Claude to introduce an error of that type. Claude is allowed to reject the request if the error cannot reasonably be applied.

After chain generation, \cds-PVD and \cds-MSR have train/test splits of 8K/2K and 8K/1K chains, respectively. The \cds dataset is useful for both caption ranking and detailed video captioning (by taking the chains' first captions). 

\subsection{Ranking Optimization}
To optimize models on our generated chains, we turn to the Plackett-Luce formulation of Direct Preference Optimization (DPO) \cite{Rafailov2023dpo}. The standard formulation of DPO follows the Bradley-Terry preference model, which defines the likelihood that a response $y_1$ is preferred over response $y_2$ given input $x$ as 
\begin{equation}
    P(y_1 \succ y_2 \mid x) = \frac{\exp(r(x,y_1))}{\exp(r(x, y_1)) + \exp(r(x, y_2))},
\end{equation}
where $r$ denotes a scalar reward model. The Plackett-Luce model generalizes the Bradley-Terry model from pairwise rankings to full rankings. The probability of observing a ranking $y_{i_1} \succ \ldots \succ y_{i_n}$ is modeled as 
\begin{equation}
    P(y_1 \succ \ldots \succ y_n \mid x) = \prod_{i = 1}^n \frac{\exp(r(x, y_i))}{\sum_{j = i}^n \exp(r(x,y_j))}.
    \label{eq:plackett_luce}
\end{equation}
Intuitively, \cref{eq:plackett_luce} states that the probability that response $y_i$ is preferred over all lower ranked responses $y_j$ ($j \ge i$) is proportional to its exponentiated reward relative to the total reward of remaining candidates. The product is taken over all responses to ensure that each response at position $i$ ``wins'' over all subsequent ones in the ranking.

DPO starts from a KL-constrained reward maximization objective for reinforcement learning, expressing the reward model $r$ in terms of the optimal policy $\pi^*$ and a reference policy $\pi_\text{ref}$. The preference model is then reparametrized with the expression for the reward model. Using the Plackett-Luce model, one obtains
\begin{equation}
    P(y_1 \succ \ldots \succ y_n \mid x) = \prod_{i = 1}^n \frac{\exp\left(\beta \log \frac{\pi^*(y_i \mid x)}{\pi_\text{ref}(y_i \mid x)}\right)}{\sum_{j = i}^n \exp\left( \beta\log \frac{\pi^*(y_j \mid x)}{\pi_\text{ref}(y_j \mid x)}\right)}.
    \label{eq:plackett_luce_dpo}
\end{equation}
Replacing $\pi^*$ with the parametrized policy $\pi_\theta$ lets us take the negative log-likelihood of \cref{eq:plackett_luce_dpo} as our loss.

%% file: sec/experiments.tex
\section{Experiments}
\label{sec:experiments}
To assess the effectiveness of our caption ranking objective, we use VLMs from two state-of-the-art model families, \textit{PerceptionLM-1B} (PLM)~\cite{cho2025perceptionlm} and \textit{Qwen2.5-VL-3B}~\cite{bai2025qwen2}, and train them on our \cds dataset. We compare the pretrained models prompted zero-shot to models trained with 1) binary, Bradley-Terry DPO trained using the top-two captions of each chain; 2) Multi-Preference Optimization (MPO)~\cite{gupta2024multi}, a generalization of DPO that can utilize the entire chain by setting the top caption as the winning one and the rest as losing ones, and 3) our proposed ranking optimization, where we use Plackett-Luce DPO to learn to rank the entire chain. Additionally, we evaluate different objectives utilizing our ordered caption chains and compare them to our Ranking objective. We then evaluate all models on video captioning and multiple choice question answering tasks. 

Unless otherwise stated, models are trained for 1K steps with a learning rate of 1e-6 on a benchmark's corresponding training set. We use a batch size of 8 and the AdamW \cite{Loshchilov2017adamw} optimizer. Models are trained end-to-end with LoRA adapters \cite{Hu2021lora}. DPO, MPO, and ranking models use $\beta = .3$ and include a next-token-prediction loss with weight .1 to prevent generation collapse \cite{dubey2024llama3, pal2024dpop}. We fine-tune models using our data generation approach and traditional approaches to show that our caption degradation method achieves strong performance, and that ranking objectives enabled by our data provides additional gains. We run experiments on a single node with 8 A100 GPUs.
\begin{table}[b]
    \small
    \centering
    \caption{Comparison of our proposed \textit{Caption Degradation} method to other on- and off-policy DPO data generation methods with the PLM model. Captions are evaluated by LLM-as-a-Judge along Relevance (\textit{Rel.}), Descriptiveness (\textit{Descr.}), Temporal Consistency (\textit{Temp C.}), and Fluency (\textit{Flu.}) on PE-Video.}
    \label{tab:data_gen}
    \resizebox{.9\linewidth}{!}{
        \begin{tblr}{
            colspec = {lccccc},
            rowsep = 0pt,
            row{1} = {bg=gray!8}
        }
            \toprule
            \textbf{Method} & \textbf{On-Policy} & \textit{Rel.} & \textit{Descr.} & \textit{Temp C.} & \textit{Flu.}  \\
            \midrule
            
            Error Injection (MPO) & \ding{55} & 4.90 & 4.62 & 5.38 & 8.46 \\
            
            Ours - Caption Degradation (Rank) & \ding{55} & \textbf{5.27} & \textbf{4.92} & \textbf{5.75} & 8.47 \\
            Response Improvement & $\checkmark$ & 5.07 & \textbf{4.92} & 5.32 & 7.61 \\
            Response Comparison & $\checkmark$ & 5.12 & 4.71 & 5.47 & \textbf{8.60} \\
            RLAIF-V & $\checkmark$ & 5.13 & 4.83 & 5.52 & 8.57 \\
            \bottomrule
        \end{tblr}
    }
\end{table}

\vspace{-0.3cm}
\subsection{Video Detailed Captioning}
The video detailed captioning task requires a model to produce a detailed description of a video, going beyond key objects, actors, and events to discuss fine details \cite{Chai2024AuroraCapEP}. Detailed captioning is an effective way to assess models' visual understanding---by requiring comprehensive descriptions, the task reveals how the model interprets visual inputs and how effectively it can communicate that understanding. Video detailed captioning introduces an extra layer of difficulty compared to image captioning by adding a temporal dimension.

\subsubsection{Video Detailed Captioning on \cds}
Motivated by existing work on caption evaluation \cite{Matsuda2025vela}, we define the following axes along which we evaluate generated captions.

\begin{itemize}
    \item \textbf{Relevance}: The caption should contain accurate and relevant statements, avoiding fabrications or errors.
    \item \textbf{Descriptiveness}: The caption should be comprehensive, capturing key objects, attributes, details, and events. 
    \item \textbf{Temporal Consistency}: Events should be described as occurring in the correct order.
    \item \textbf{Fluency}: Descriptions should be coherent and fluent. 
\end{itemize}

We perform a reference-based LLM-as-a-judge \cite{Gu2024ASO} evaluation where we provide the LLM judge with the predicted and reference captions, prompting it to score each of the above metrics from one to ten.

\paragraph{Evaluation of Our Data Generation Method.}
Our proposed method differs from standard DPO in two ways: 1) data generation, by generating chains through caption degradation, and 2) in its use of a ranking objective, enabled by the generation of ranked chains. We start by comparing our \textit{Caption Degradation} method to other automated DPO data generation pipelines. Our baselines represent the most common procedures to obtain preference pairs at scale \cite{yang2025onpolicydpo}: 1) \textit{Error Injection} into ground truth responses (without repeated degradation), 2) \textit{Response Improvement} by a teacher LLM, 3) Pairwise \textit{Response Comparison} by a teacher LLM, and 4) \textit{RLAIF-V}, a claim extraction, assessment, and subsequent response ranking paradigm \cite{yu2025rlaifv} (details in Appendix). For this comparison, we use GPT-OSS-20B \cite{IntroducingGPTOSS2025} for data generation and evaluation on PE-Video as it maintains strong performance while being more economical than Claude Sonnet 3.7 for generating multiple datasets.

\cref{tab:data_gen} shows the results of training PLM on each dataset. Our caption degradation method obtains the highest scores of all on- and off-policy methods as measured by \textit{Relevance}, \textit{Descriptiveness}, and \textit{Temporal Consistency}, while maintaining competitive \textit{Fluency}. Our approach's simplicity as an extension of error injection, coupled with its strong performance and scalability, demonstrate its practicality.
\paragraph{Evaluation of Our Ranking Objective.} Having demonstrated the effectiveness of our caption degradation method for preference data generation, we use Claude Sonnet 3.7 to generate the Ranked Caption Chain (\cds) dataset as described in \cref{sec:rcc}. To ensure equivalent training and evaluation regimes, all subsequent results are obtained by training on \cds and evaluating using Sonnet 3.7 as the LLM judge. 
\begin{table*}
    \small
    \centering
    \caption{Captioning results on PE-Video and MSR-VTT splits of \cds. Models trained with the ranking objective perform best across the board, with DPO-based methods outperforming SFT. We note that MPO and DPO benefit from captions generated with our Caption Degradation method as they are fine-tuned on it.}
    \resizebox{\linewidth}{!}{
        \begin{tblr}{
            colspec = {l *{8}{Q[c]} *{8}{Q[c]}},
            colsep = 3pt,
            rowsep = 1pt,
            cell{1}{2} = {c=8}{c, bg=blue!8, font=\bfseries\itshape},
            cell{1}{10} = {c=8}{c, bg=orange!8, font=\bfseries\itshape},
            row{2} = {bg=gray!8, font=\bfseries},
            row{3} = {bg=gray!8, font=\itshape},
        }
            \toprule
            \SetCell[r=3]{l, m, bg=gray!8, font=\bfseries} Method & PerceptionLM & & & & & & & & Qwen2.5-VL & & & & & & & \\
            \cmidrule[lr]{2-9}\cmidrule[lr]{10-17}
            & \SetCell[c=4]{c} \textsc{MSR-VTT} & & & & \SetCell[c=4]{c} \textsc{PE-Video} & & & & \SetCell[c=4]{c} \textsc{MSR-VTT} & & & & \SetCell[c=4]{c} \textsc{PE-Video} & & & \\
            \cmidrule[lr]{2-5}\cmidrule[lr]{6-9}\cmidrule[lr]{10-13}\cmidrule[lr]{14-17}
            & Rel. & Descr. & Temp C. & Flu. & Rel. & Descr. & Temp C. & Flu. & Rel. & Descr. & Temp C. & Flu. & Rel. & Descr. & Temp C. & Flu. \\
            \midrule
            Pretrained & 6.29 & 5.49 & 6.18 & 7.19 & 6.72 & 6.04 & 5.82 & 6.98 & 6.73 & 6.14 & 6.61 & 8.69 & 6.74 & 6.12 & 6.63 & 8.69 \\
            SFT        & 6.44 & 6.37 & 6.98 & 8.54 & 6.35 & 5.60 & 6.07 & 8.15 & 7.13 & 6.83 & 7.76 & 9.14 & 7.31 & 7.24 & 7.64 & 9.35 \\
            DPO        & 7.29 & 7.34 & 8.02 & 9.33 & 7.57 & 7.59 & 7.78 & 9.43 & 7.12 & 6.86 & 7.77 & 9.13 & 7.33 & 7.25 & 7.60 & 9.36 \\
            MPO        & 7.31 & 7.38 & 8.08 & 9.32 & 7.57 & 7.59 & 7.80 & 9.42 & 7.11 & 6.89 & 7.84 & \textbf{9.15} & 7.35 & 7.27 & 7.66 & 9.34 \\
            Rank       & \textbf{7.38} & \textbf{7.46} & \textbf{8.11} & \textbf{9.37} & \textbf{7.67} & \textbf{7.72} & \textbf{7.93} & \textbf{9.49} & \textbf{7.21} & \textbf{6.93} & \textbf{7.86} & \textbf{9.15} & \textbf{7.46} & \textbf{7.40} & \textbf{7.75} & \textbf{9.40} \\
            \bottomrule
        \end{tblr}
    }
    \label{tab:captioning}
    \vspace{-1em}
\end{table*}

\cref{tab:captioning} shows the results on the \cds evaluation splits. We find that \textbf{performance generally correlates with data utilization}. Ranking, which uses the whole chain and its ordering, performs best across the board. MPO also uses our ordered caption chain but discards the ordering, performing worse than Ranking on all metrics. DPO uses only the top two captions of our ordered caption chains, and SFT has no negatives to compare against, with PLM's variant performing far worse than the DPO variants. Fluency has the smallest increase amongst the DPO variants, suggesting our data generation process successfully targets visual details while retaining caption fluency.

To ensure that evaluations are not biased due to use of the same LLM for data generation and evaluation, we recompute the scores of \cref{tab:captioning} using GPT-OSS-20B, then compute the Spearman correlation between GPT-OSS' and Claude 3.7 Sonnet's captioning scores. Outside of a Fluency correlation of .45, we observe moderately strong correlations of .72 for Relevance, .62 for Descriptiveness, and .65 for Temporal Consistency. This indicates that the judges tend to agree when a caption deserves a higher or lower score.  Full results are in the Appendix.

\begin{wraptable}{r}{0.42\columnwidth}
\vspace{-4em}
\centering
\setlength{\tabcolsep}{3pt}

\caption{Head-to-head human evaluation of PLM on \textsc{Rcc}-PVD. Ranking objective combined with ordered caption chains outperform DPO and MPO 20\% and 21\%. On the other hand, our ordered caption chain data improves the MPO and DPO baselines by 7\% and 5\%.}
\resizebox{\linewidth}{!}{
    \begin{tabular}{l cc cc}
    \toprule
    & \multicolumn{2}{c}{\textbf{Cap. Chains}} & \multicolumn{2}{c}{\textbf{Err. Inj.}} \\
    \cmidrule(lr){2-3} \cmidrule(lr){4-5}
    & \textit{Rank.} & \textit{Other} & \textit{Rank.} & \textit{Other} \\
    \midrule
    
    vs.\ MPO & \cellcolor{blue!8}\textbf{53\%} & 47\% & \cellcolor{blue!8}\textbf{60\%} & 40\% \\
    vs.\ DPO & \cellcolor{blue!8}\textbf{56\%} & 44\% & \cellcolor{blue!8}\textbf{61\%} & 39\% \\
    \bottomrule
    \end{tabular}
}
\label{tab:human_eval}
\end{wraptable}

\paragraph{Human Evaluation.} We also perform a \textbf{head-to-head human evaluation} of PLM trained on captions from \textsc{Rcc}-PVD with Ranking, MPO, and DPO (Tab.~\ref{tab:human_eval}). We sample 100 videos from the evaluation set for each comparison. When all methods are trained using our proposed caption chains, Ranking is preferred over MPO 53\% of the time and over DPO 56\% of the time. When we only train the Ranking model with caption chains and MPO and DPO are trained using error injection, Ranking is preferred over MPO 60\% of the time and over DPO 61\% of the time. These results show that our data generation method (caption chains) combined with the ranking objective is preferred $ \approx 20\%$ of the time more than MPO and DPO. In the rest of the experiments section, we fine-tune the baselines using only our data generation approach (Ordered Caption Chains) as it achieves the best results for both baselines and our Ranking objective.

\begin{table}[!h]
    \vspace{-1em}
    \small
    \centering
    \caption{Results on the VDC and ARGUS video captioning benchmarks. Ranking-based models show strong performance from both families. MPO and DPO benefit from our data generation approach (ordered captions) as they are fine-tuned on it.}
    \resizebox{0.9\textwidth}{!}{
    \begin{tblr}{
        colspec = {l *{4}{c} *{4}{c}},
        colsep = 6pt,
        rowsep = 1pt,
        row{1} = {font=\bfseries\itshape},
        row{2} = {font=\bfseries, bg=gray!8},
        row{3} = {font=\itshape, bg=gray!8},
        cell{1}{1} = {c=5}{c, bg=blue!8},
        cell{1}{6} = {c=4}{c, bg=orange!8},
        cell{2}{1} = {r=2}{m},
        cell{2}{2} = {c=2}{c},
        cell{2}{4} = {c=2}{c},
        cell{2}{6} = {c=2}{c},
        cell{2}{8} = {c=2}{c},
    }
        \toprule
        PerceptionLM & & & & & Qwen2.5-VL & & & \\
        \midrule
        Method & \SetCell[c=2]{c} \textsc{VDC} & & \SetCell[c=2]{c} \textsc{Argus} & & \SetCell[c=2]{c} \textsc{VDC} & & \SetCell[c=2]{c} \textsc{Argus} & \\
        \cmidrule[lr]{2-3} \cmidrule[lr]{4-5} \cmidrule[lr]{6-7} \cmidrule[lr]{8-9}
        & Score & Accuracy & Cost-H $(\downarrow)$ & Cost-O $(\downarrow)$ & Score & Accuracy & Cost-H $(\downarrow)$ & Cost-O $(\downarrow)$ \\
        \midrule
        Pretrained & 2.01 & .382 & .679 & .895 & 2.30 & .443 & \textbf{.664} & .820 \\
        SFT & 1.95 & .373 & .712 & .875 & 2.33 & .453 & .703 & .786 \\
        DPO & 2.26 & .436 & .570 & .806 & 2.34 & .454 & .683 & .786 \\
        MPO & 2.27 & .438 & .572 & .801 & \textbf{2.35} & \textbf{.455} & .684 & \textbf{.772} \\
        Rank & \textbf{2.30} & \textbf{.445} & \textbf{.562} & \textbf{.786} & 2.34 & .453 & .681 & \textbf{.772} \\
        \bottomrule
    \end{tblr}
    }
    \label{tab:captioning_external}
    \vspace{-3em}
\end{table}
\subsubsection{Video Detailed Captioning on VDC and ARGUS}
Next, we evaluate models trained on RCC-PVD and RCC-MSR with existing video captioning benchmarks. The VDC \cite{Chai2024AuroraCapEP} and ARGUS \cite{Rawal2025ARGUSHA} benchmarks both focus on detailed video captions but evaluate them in different ways. VDC takes a polling approach, with a set of questions for each caption used for evaluation. An external LLM is used to answer these questions predicated on the evaluated model's predicted caption and evaluate the predicted caption's quality. These correctness and quality scores are averaged to produce the final VDCscore. 

ARGUS evaluates captions on \textit{hallucination} (Cost-H), whether the the predicted caption contains untrue statements, and \textit{omission} (Cost-O), whether information in the reference captions is contained in the prediction. These costs are computed by splitting the predicted and reference captions into sentences, determining whether each predicted sentence is entailed by a reference sentence, and combining an alignment cost between the predicted and reference sentences with an order penalty if events are described out of order. We use Claude 3.7 Sonnet to perform the natural language entailment.

Results on these benchmarks are shown in \cref{tab:captioning_external}. The ranking variant of PerceptionLM performs better than the non-ranking baselines on both VDC and ARGUS. The results with Qwen2.5 are mixed, with all variants performing comparably on VDC. The ranking model performs best on ARGUS (the pretrained model obtains low hallucination and high omission costs by generating short responses). The consistently strong performance of the ranking-trained models demonstrates the utility of our approach.
\vspace{-0.1cm}
\subsection{Multiple Choice Question Answering}
To assess whether our caption ranking method improves models on tasks beyond caption generation, we evaluate our models on multiple-choice question answering. 

\vspace{-0.3cm}
\subsubsection{Long-form Multiple Choice Question Answering}
\label{sec:lf-mcqa}
We define \textit{long-form multiple choice question answering} (LF-MCQA) as a task where the model is asked a question about a provided input (text or visual) and is asked to select from multiple long, detailed answer choices. Such answer choices may be descriptive, having the model select the answer choice which best describes the input; analytical, prompting the model to interpret the input; and inferential, pushing the model to draw conclusions based on the input.

As our focus is on video understanding, we create an evluation benchmark by starting from PE-Video's \cite{Bolya2025PerceptionET} test set and sampling 2K human-annotated video descriptions. We provide each description to Claude and prompt it to generate a question with five answer choices, each of which is a rich paragraph. Manual inspection shows that the questions and answer choices are challenging and of high quality. One such example is shown below, with full examples provided in the Appendix. 
\begin{quote}
{\scriptsize
\textbf{Question:} Based on the aerial footage of the construction site, 
what conclusions can be drawn about the nature of the development, its 
planning context, and potential environmental and community impacts?

\begin{enumerate}[label=\textbf{\Alph*.}, itemsep=0.3em, leftmargin=*]
  \item Typical urban infill development integrated with surrounding buildings \ldots
  \item Large-scale residential development expanding an existing area and coordinated with current roads/buildings \textbf{(Correct)}
  \item Emergency post-disaster housing with temporary structures and hastily built access roads \ldots
  \item Environmentally focused development with solar panels and rehabilitated green spaces \ldots
  \item Government-subsidized affordable housing with uniform structures and separation from higher-value properties \ldots
\end{enumerate}
}
\end{quote}
To evaluate a model on our LF-MCQA dataset, we provide them with a video, question, and all five answer choices. Answers are predicted by selecting the answer choice whose letter has the highest log probability (e.g. the log probability of the string ``A'').

\cref{tab:lf-mcqa} shows the results. The SFT variants of both models perform no better than the pretrained ones. DPO, MPO, and ranking models all obtain higher performance than the SFT variants, with the ranking model performing best. These results suggest that caption ranking not only allows the model to better generate its own captions, but also trains it to better evaluate the phrases in the prompt---in this case, each individual answer choice. 
\vspace{-0.6cm}
\subsubsection{Transforming Captions for Question Answering}
\label{sec:transforming_captions}
In \cref{sec:lf-mcqa} we showed that our caption ranking approach improved long-form multiple choice question answering, a task it was not directly optimized for. Here, we ask the following: can we \textit{rephrase our caption chains as question-answer chains} to benefit other tasks?
\begin{table}[b]
    \centering
    \begin{minipage}[t]{0.48\textwidth}
        \centering
        \caption{Results on our Long-form Multiple Choice Question Answering benchmark. Models trained using the \cds-PVD caption chains improve on this task despite not being trained for it.}
        \resizebox{\textwidth}{!}{
            \begin{tblr}{
                colspec = {l c l c},
                row{1} = {bg=gray!8},
                rowsep = 1pt
            }
                \toprule
                \textbf{PerceptionLM} & \textit{Acc.} & \textbf{Qwen2.5-VL} & \textit{Acc.} \\
                \midrule
                Pretrained & .618 & Pretrained & .820 \\
                SFT & .619 & SFT & .819 \\
                DPO & \textbf{.644} & DPO & .826 \\
                MPO & .642 & MPO & .825 \\
                Rank & \textbf{.644} & Rank & \textbf{.832} \\
                \bottomrule
            \end{tblr}
        }
        \label{tab:lf-mcqa}
    \end{minipage}
    \hfill
    \begin{minipage}[t]{0.48\textwidth}
        \centering
        \small
        \caption{Effect of different ranking formulations during training. MCQ indicates the model is trained to rank multiple-choice answers, YNQ that it is trained to rank yes/no question-answer pairs, and Cap that it is trained with normal caption ranking. Numbers are average accuracy on TempCompass Caption Matching.}
        \resizebox{\textwidth}{!}{
        \begin{tblr}{
            colspec = {lccc},
            row{1} = {font=\bfseries, bg=gray!8},
        }
            \toprule
            PerceptionLM & MCQ & MCQ/YNQ & MCQ/YNQ/Cap \\
            \midrule
            Rank & 62.61 & 64.27 & 64.94 \\
            \midrule
            Pretrained &  & 43.25  &  \\
            \bottomrule
        \end{tblr}
        }
        \label{tab:tempcompass}
    \end{minipage}
\end{table}
To test this possibility, we turn to the Something-Something v2 (SSv2) \cite{goyal2017ssv2} dataset. We generate caption chains by repeatedly modifying a caption's object or action, then have an LLM reformat the captions into questions where the varying object/actions in the chain serve as the answer choices of decreasing quality. We train models on these MCQ chains by having them rank the answer choice strings (e.g. ``(a)'', ``(b)'') given a question listing all the answer choices.

For yes/no question (YNQ) chains, we similarly provide an LLM with the caption chain and prompt it to reformat each caption into a yes/no question focused on the varying objects/actions in the chain (e.g. ``Is there an \texttt{[object]} in the video?''. Unlike with caption or multiple choice ranking which varies the response for a fixed prompt, here we vary the prompt for the fixed response ``no''. Intuitively, as the questions become increasingly erroneous, the model should become more confident in answering ``no''. We hence reverse the order of the chain and rank the likelihood of the response ``no'' for each question in the chain. 

We evaluate models trained on these caption, MCQ, and yes/no question chains on the TempCompass Caption Matching benchmark. For this task, a model must generate the answer choice that corresponds to the best-matching caption for a video. We assess the effectiveness of reformatting caption chains into question-answer chains by training one model with only MCQ chains, one model with data split between MCQ and YNQ chains, and a third model with data split between MCQ, YNQ, and caption chains. The results in \cref{tab:tempcompass} show that even with a small compute budget of 500 optimization steps, rephrasing the caption chains as question-answering chains provides data diversity that improves performance on the downstream task.
\begin{table}[b]
    \vspace{-1em}
    \centering
    \small
    \caption{Performance of Qwen2.5-VL with a frozen (\ding{55}) or fine-tuned ($\checkmark$) vision encoder on PE-Video. Fine-tuning the Vision Encoder is critical for our Ranking objective.}
    \resizebox{0.75\linewidth}{!}{
        \begin{tblr}{
            colspec={l *{6}{c}},
            row{1}={font=\bfseries, bg=gray!8},
            row{2}={font=\itshape, bg=gray!8},
            rowsep=1pt,
            hline{2}={2-7}{},
        }
            \toprule
            \SetCell[r=2]{l} \textbf{Metric}
                & \SetCell[c=2]{c} SFT & 
                & \SetCell[c=2]{c} DPO & 
                & \SetCell[c=2]{c} Rank & \\
                & \ding{55} & $\checkmark$
                & \ding{55} & $\checkmark$
                & \ding{55} & $\checkmark$ \\
            \midrule
            Relevance    & 7.31 & 7.34 & 7.33 & 7.35 & 7.34 & 7.46 \\
            Descriptiveness  & 7.24 & 7.25 & 7.23 & 7.25 & 7.25 & 7.40 \\
            Temporal Consistency & 7.64 & 7.62 & 7.61 & 7.60 & 7.64 & 7.75 \\
            Fluency   & 9.35 & 9.35 & 9.34 & 9.36 & 9.36 & 9.40 \\
            \bottomrule
        \end{tblr}
    }
    \label{tab:frozen_ve}
\end{table}
\vspace{-0.3cm}
\subsection{Design Choices}
\subsubsection{Finetuning the Vision Encoder}
In \cref{tab:frozen_ve} we examine the impact of freezing the vision encoder while finetuning Qwen2.5-VL on \cds-PVD. The results reveal a notable difference between training paradigms: while SFT and DPO exhibit marginal performance variations between frozen and trainable vision encoder configurations, the ranking method achieves substantial gains when the vision encoder is optimized jointly with the language model. This disparity suggests that \textbf{finetuning the language model alone is insufficient} to rank the captions. End-to-end training enables the vision encoder to learn the fine-grained visual representations necessary to  discriminate between subtle changes in the caption chains. These findings suggest that the common practice of freezing the vision encoder during DPO is suboptimal for vision-heavy tasks, preventing the vision encoder from benefiting from the contrastive learning induced by DPO.
\vspace{-0.7cm}
\subsubsection{Impact of Chain Length}
\begin{figure}[t]
    \centering
    \includegraphics[width=.70\linewidth]{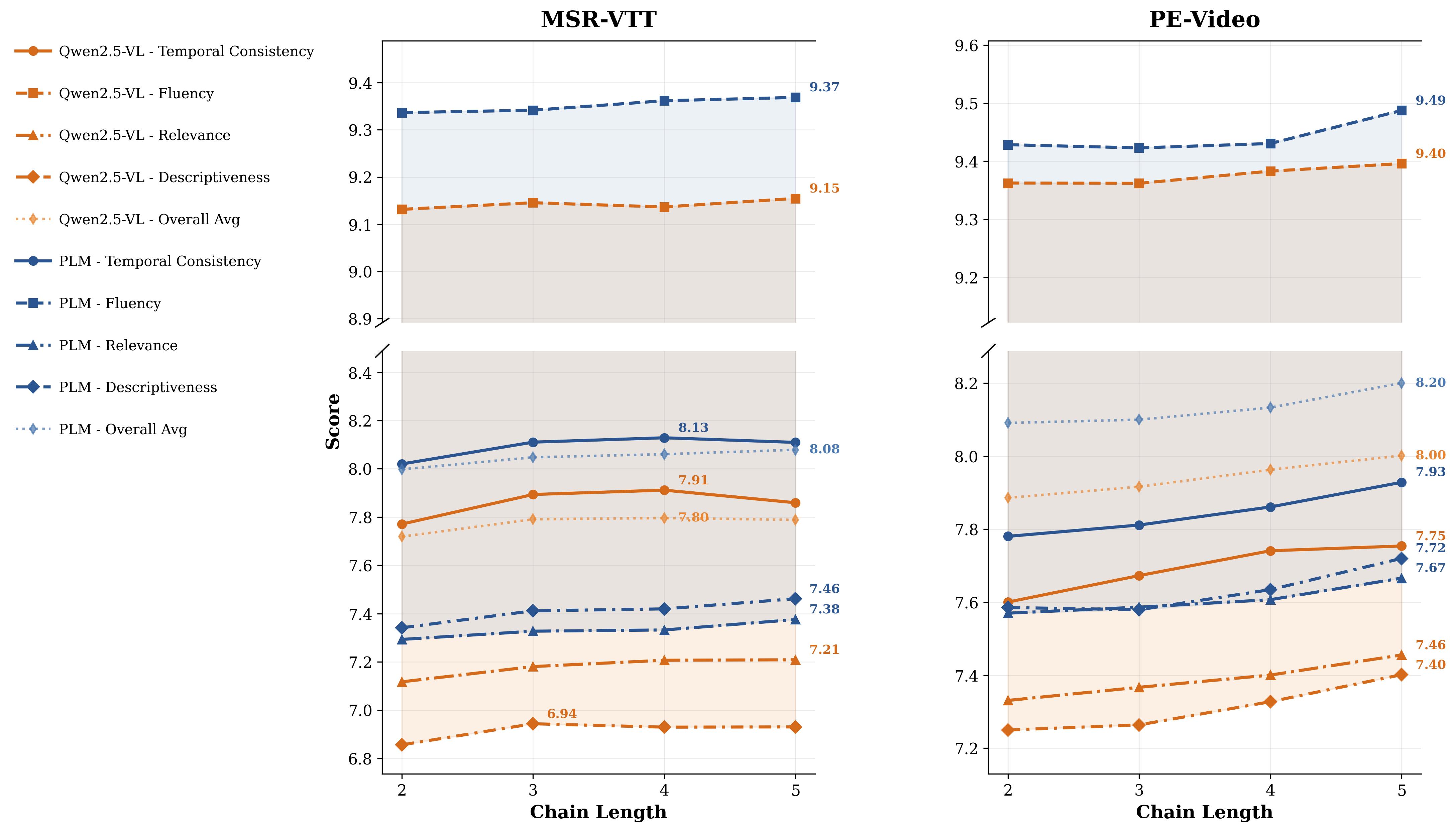}
    \caption{Performance of Qwen2.5-VL and PLM w.r.t number of captions utilized in the chain. Model performance generally increases with chain length.}
    \label{fig:perf_by_chain_len}
    \vspace{-1em}
\end{figure}
\begin{table}[b]
    \centering
    \caption{Comparison between models trained with different losses on \cds-PVD.}
    \label{tab:loss_comparison}
    \resizebox{0.9\textwidth}{!}{
        \begin{tblr}{
            colspec=lcccccccc,
            rowsep=1pt,
            row{1-2}={bg=gray!9},
            row{2}={font=\itshape},
            hline{2}={2-5}{leftpos=-1,rightpos=-1,endpos},
            hline{2}={6-9}{leftpos=-1,rightpos=-1,endpos},
            cell{1}{2}={bg=blue!8},
            cell{1}{6}={bg=orange!8},
        }
            \toprule
            \SetCell[r=2]{} \textbf{Loss} 
                & \SetCell[c=4]{} \textbf{PerceptionLM} & & & & 
                \SetCell[c=4]{} \textbf{Qwen2.5-VL} & & & \\
            & Rel. & Descr. & Temp C. & Flu. & Rel. & Descr. & Temp C. & Flu. \\
            \midrule
            Hinge  & 7.58 & \textbf{7.72} & 7.88 & 9.47 & 7.34 & 7.36 & \textbf{7.79} & 9.38 \\
            Plackett-Luce DPO & \textbf{7.67} & \textbf{7.72} & \textbf{7.93} & \textbf{9.49} & \textbf{7.46} & \textbf{7.40} & 7.75 & \textbf{9.40} \\
            \bottomrule
        \end{tblr}
    }
\end{table}
We investigate the impact of chain length on model performance in \cref{fig:perf_by_chain_len}. The results demonstrate that increasing chain length consistently yields improved performance across both Qwen2.5-VL and PLM architectures. Further exploration of the scalability limits with extended chain lengths remains a promising direction for future research.

\vspace{-0.6cm}
\subsubsection{Choice of Ranking Loss}

For a fixed set of responses, the Plackett-Luce loss encourages larger probability ratios between higher-ranked responses and all lower-ranked ones. While this appears to exert downward pressure on lower ranks, this may not occur in practice, even for binary DPO \cite{pal2024dpop}. Because this loss places no constraint on the trajectory of individual likelihoods, an undesirable response may increase in probability. Indeed, we found that a next-token prediction (NTP) loss was necessary to prevent VLMs' generation abilities from collapsing.

These considerations motivate examining the margin-free hinge loss as an alternative, since it penalizes the model only when a lower-ranked response receives a higher score than a higher-ranked one. As it acts only on score inversions, it avoids the margin-widening pressure exerted by the Plackett-Luce loss.

We therefore train models with hinge loss and NTP loss and compare them to our models trained with Plackett-Luce + NTP. Given log probabilities $s_i$ corresponding to responses ordered as $y_1 \succ \ldots \succ y_n$, we compute the margin-free pairwise hinge loss as
\begin{equation}
    \mathcal{L} =\frac{2}{n(n - 1)}\sum_{i = 1}^{n - 1} \sum_{j = i + 1}^n \max(0, s_j - s_i).
\end{equation}
\cref{tab:loss_comparison} shows the results. Despite the conceptual appeal of limiting the loss to inversions, the Plackett-Luce models generally perform better in practice.

\vspace{-0.5cm}
\subsubsection{Conditioned vs. Unconditioned Caption Mutation}

To build the \cds dataset, we uniformly sampled error types and conditioned the LLM's prompt on the sampled error. An alternative could have been to allow the language model to freely introduce errors of its own choosing, without any error type conditioning. To ablate this design choice, we regenerated caption chains on PE-Video without error type conditioning. The PerceptionLM model trained on the error-conditioned chains achieved significantly higher scores across all four LLM-as-a-judge metrics (table in Appendix).  

\definecolor{hlgreen}{RGB}{200,245,200}
\definecolor{hlred}{RGB}{255,210,210}
\newcommand{\hlg}[1]{\sethlcolor{hlgreen}\hl{#1}}
\newcommand{\hlr}[1]{\sethlcolor{hlred}\hl{#1}}

\begin{figure*}[t]
    \centering
    \begin{tikzpicture}[
        scale=0.92,
        every node/.style={transform shape},
        video/.style={
            rectangle,
            rounded corners=5pt,
            minimum width=2.8cm,
            minimum height=1.8cm,
            draw=blue!60,
            line width=1.5pt,
            fill=blue!10,
            drop shadow={opacity=0.3, shadow xshift=2pt, shadow yshift=-2pt}
        },
        caption/.style={
            rectangle,
            rounded corners=4pt,
            text width=5.2cm,
            draw=black!50,
            line width=0.8pt,
            fill=white,
            align=justify,
            font=\tiny,
            inner sep=4pt,
            drop shadow={opacity=0.2, shadow xshift=1.5pt, shadow yshift=-1.5pt}
        },
        label/.style={
            font=\sffamily\bfseries\small,
            anchor=south,
            text=black,
            align=center
        },
        scale=.8,
        transform shape
    ]

    \node[video] (v1) at (0,0) {\includegraphics[width=2.8cm, height=1.8cm]{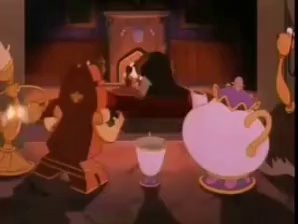}};
    \node[caption, anchor=west] (sft1) at ([xshift=0.25cm]v1.east) {A sequence from the Disney movie `The Little Mermaid' showing the interior of a castle with a fireplace. The scene begins with a window view of a snowy landscape outside where a character in a purple outfit is sitting by a tree. Inside the castle, a teapot with a crown on its lid sits on a table next to a golden candlestick holder. The sequence appears to be from a special effects sequence or a dramatic scene from the movie, capturing a dramatic confrontation between the creature and the woman.};
    \node[caption, anchor=west] (our1) at ([xshift=0.25cm]sft1.east) {A sequence from an animated Disney film showing a snowy winter scene viewed through a window. The sequence begins with a \hlg{golden candlestick holder with burning flames} on the left side of the window. As the scene progresses, a \hlg{teapot with a crown on its lid} appears in the center, accompanied by a \hlg{clock figure to its right}. The characters then transition to a \hlg{red room where a woman in a pink dress stands near a fireplace}. The sequence concludes with the teapot and clock figure reappearing on the window sill, with the burning candlestick holder still visible on the left.};

    \path let \p1=(v1.south), \p2=(sft1.south), \p3=(our1.south) in
        coordinate (row1bot) at (0, {min(\y1, min(\y2, \y3))});

    \node[video, anchor=north] (v2) at ([yshift=-0.5cm]row1bot -| v1) {\includegraphics[width=2.8cm, height=1.8cm]{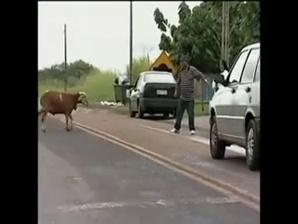}};
    \node[caption, anchor=north west] (sft2) at ([yshift=-0.5cm]row1bot -| sft1.west) {A series of frames show a brown goat on a road interacting with a woman in a white top and blue pants. The goat approaches the woman sitting on the road, raises its head, and is fed by the woman. A man in a striped shirt walks towards them, jumps onto the woman, and then exits the frame. Vehicles are parked along the road, and trees are visible in the background. The sequence captures the interaction between the goat, the woman, and the man, set against a rural or suburban road environment.};
    \node[caption, anchor=north west] (our2) at ([yshift=-0.5cm]row1bot -| our1.west) {A sequence of images showing a goat on a street \hlg{attempting to cross a road}. The goat, with its \hlg{brown coat and white markings}, is seen walking along the road, seemingly unaware of the approaching traffic. A woman in a white top and blue pants sits on the roadside. A man in a striped shirt and dark pants \hlg{approaches the scene, attempting to interact with the goat}. The goat continues to walk along the road, seemingly undeterred by the man's presence. The setting appears to be a suburban or rural area with parked cars and trees lining the road.};

    \begin{scope}[on background layer]
        \node[fit=(v1)(v2), fill=blue!15, rounded corners=3pt, inner sep=4pt] (bg1) {};
        \node[fit=(sft1)(sft2), fill=purple!15, rounded corners=3pt, inner sep=4pt] (bg2) {};
        \node[fit=(our1)(our2), fill=red!15, rounded corners=3pt, inner sep=4pt] (bg3) {};
    \end{scope}

    \node[label] at ([yshift=6pt]bg1.north) {Video};
    \node[label] at ([yshift=6pt]bg2.north) {SFT};
    \node[label] at ([yshift=6pt]bg3.north) {Ranking};

    \foreach \v in {v1,v2} {
        \begin{scope}[shift={(\v.center)}]
            \fill[white, opacity=0.7] (0,0) circle (0.25cm);
            \fill[blue!70] (-0.08,-0.12) -- (-0.08,0.12) -- (0.16,0) -- cycle;
        \end{scope}
    }
    
    \end{tikzpicture}
    \caption{Qualitative comparison of our Ranking model vs.\ an SFT baseline. \sethlcolor{hlgreen}\hl{Green} highlights video-grounded descriptions. The Ranking model captures finer details better than SFT. See Appendix for more in-depth comparison and examples.}
    \label{fig:qualitative}
\end{figure*}

%% file: sec/conclusion.tex
\vspace{-0.2cm}
\section{Conclusion}
\vspace{-0.2cm}

In this work, we addressed limitations of binary preference optimization for vision language models by introducing ranking optimization as a more nuanced alternative. Through experiments on detailed video captioning and VQA, we showed that ranking methods combined with our ordered captions generation consistently outperform DPO baselines, particularly for detailed content. Our caption degradation technique enables creation of totally ordered training data that better captures the spectrum of response quality. Critically, ranking optimization requires vision encoder finetuning to achieve meaningful improvements, revealing that preference optimization in VLMs extends beyond simple language reweighting. Our results suggest that future work should embrace graded preference structures and jointly optimize both vision and language components.

%% file: references.bib
@String(CVPR= {IEEE Conf. Comput. Vis. Pattern Recog.})

@String(CVPR  = {CVPR})

@inproceedings{liu2024tempcompass,
  title={TempCompass: Do Video LLMs Really Understand Videos?},
  author={Yuanxin Liu and Shicheng Li and Yi Liu and Yuxiang Wang and Shuhuai Ren and Lei Li and Sishuo Chen and Xu Sun and Lu Hou},
  booktitle={Annual Meeting of the Association for Computational Linguistics},
  year={2024},
  url={https://api.semanticscholar.org/CorpusID:268201547}
}

@article{Tong2024eyeswideshut,
  title={Eyes Wide Shut? Exploring the Visual Shortcomings of Multimodal LLMs},
  author={Shengbang Tong and Zhuang Liu and Yuexiang Zhai and Yi Ma and Yann LeCun and Saining Xie},
  journal={2024 IEEE/CVF Conference on Computer Vision and Pattern Recognition (CVPR)},
  year={2024},
  pages={9568-9578},
  url={https://api.semanticscholar.org/CorpusID:266976992}
}

@article{xie2024v,
  title={V-dpo: Mitigating hallucination in large vision language models via vision-guided direct preference optimization},
  author={Xie, Yuxi and Li, Guanzhen and Xu, Xiao and Kan, Min-Yen},
  journal={arXiv preprint arXiv:2411.02712},
  year={2024}
}

@inproceedings{zhang2025direct,
  title={Direct preference optimization of video large multimodal models from language model reward},
  author={Zhang, Ruohong and Gui, Liangke and Sun, Zhiqing and Feng, Yihao and Xu, Keyang and Zhang, Yuanhan and Fu, Di and Li, Chunyuan and Hauptmann, Alexander G and Bisk, Yonatan and others},
  booktitle={Proceedings of the 2025 Conference of the Nations of the Americas Chapter of the Association for Computational Linguistics: Human Language Technologies (Volume 1: Long Papers)},
  pages={694--717},
  year={2025}
}

@inproceedings{he2020momentum,
  title={Momentum contrast for unsupervised visual representation learning},
  author={He, Kaiming and Fan, Haoqi and Wu, Yuxin and Xie, Saining and Girshick, Ross},
  booktitle={Proceedings of the IEEE/CVF conference on computer vision and pattern recognition},
  pages={9729--9738},
  year={2020}
}

@inproceedings{chen2020simple,
  title={A simple framework for contrastive learning of visual representations},
  author={Chen, Ting and Kornblith, Simon and Norouzi, Mohammad and Hinton, Geoffrey},
  booktitle={International conference on machine learning},
  pages={1597--1607},
  year={2020},
  organization={PmLR}
}

@article{chen2020improved,
  title={Improved baselines with momentum contrastive learning},
  author={Chen, Xinlei and Fan, Haoqi and Girshick, Ross and He, Kaiming},
  journal={arXiv preprint arXiv:2003.04297},
  year={2020}
}

@inproceedings{ouali2024clip,
  title={Clip-dpo: Vision-language models as a source of preference for fixing hallucinations in lvlms},
  author={Ouali, Yassine and Bulat, Adrian and Martinez, Brais and Tzimiropoulos, Georgios},
  booktitle={European Conference on Computer Vision},
  pages={395--413},
  year={2024},
  organization={Springer}
}

@article{Yuksekgonul2022when_and_why,
  title={When and why vision-language models behave like bags-of-words, and what to do about it?},
  author={Mert Yuksekgonul and Federico Bianchi and Pratyusha Kalluri and Dan Jurafsky and James Y. Zou},
  journal={ArXiv},
  year={2022},
  volume={abs/2210.01936},
  url={https://api.semanticscholar.org/CorpusID:252734947}
}

@article{Rafailov2023dpo,
  title={Direct Preference Optimization: Your Language Model is Secretly a Reward Model},
  author={Rafael Rafailov and Archit Sharma and Eric Mitchell and Stefano Ermon and Christopher D. Manning and Chelsea Finn},
  journal={ArXiv},
  year={2023},
  volume={abs/2305.18290},
  url={https://api.semanticscholar.org/CorpusID:258959321}
}

@article{Huang2024llm2clip,
  title={LLM2CLIP: Powerful Language Model Unlocks Richer Visual Representation},
  author={Weiquan Huang and Aoqi Wu and Yifan Yang and Xufang Luo and Yuqing Yang and Liang Hu and Qi Dai and Xiyang Dai and Dongdong Chen and Chong Luo and Lili Qiu},
  journal={ArXiv},
  year={2024},
  volume={abs/2411.04997},
  url={https://api.semanticscholar.org/CorpusID:273877534}
}

@article{Schulman2017ppo,
  title={Proximal Policy Optimization Algorithms},
  author={John Schulman and Filip Wolski and Prafulla Dhariwal and Alec Radford and Oleg Klimov},
  journal={ArXiv},
  year={2017},
  volume={abs/1707.06347},
  url={https://api.semanticscholar.org/CorpusID:28695052}
}

@article{Shao2024grpo,
  title={DeepSeekMath: Pushing the Limits of Mathematical Reasoning in Open Language Models},
  author={Zhihong Shao and Peiyi Wang and Qihao Zhu and Runxin Xu and Jun-Mei Song and Mingchuan Zhang and Y. K. Li and Yu Wu and Daya Guo},
  journal={ArXiv},
  year={2024},
  volume={abs/2402.03300},
  url={https://api.semanticscholar.org/CorpusID:267412607}
}

@inproceedings{zhou2024streaming,
  title={Streaming dense video captioning},
  author={Zhou, Xingyi and Arnab, Anurag and Buch, Shyamal and Yan, Shen and Myers, Austin and Xiong, Xuehan and Nagrani, Arsha and Schmid, Cordelia},
  booktitle={Proceedings of the IEEE/CVF Conference on Computer Vision and Pattern Recognition},
  pages={18243--18252},
  year={2024}
}

@article{gupta2024multi,
  title={Multi-Preference Optimization: Generalizing DPO via Set-Level Contrasts},
  author={Gupta, Taneesh and Madhavan, Rahul and Zhang, Xuchao and Natarajan, Nagarajan and Bansal, Chetan and Rajmohan, Saravan},
  journal={arXiv preprint arXiv:2412.04628},
  year={2024}
}

@article{poppi2026countervid,
  title={CounterVid: Counterfactual Video Generation for Mitigating Action and Temporal Hallucinations in Video-Language Models},
  author={Poppi, Tobia and Uzkent, Burak and Garg, Amanmeet and Porto, Lucas and Kessler, Garin and Yang, Yezhou and Cornia, Marcella and Baraldi, Lorenzo and Cucchiara, Rita and Schiffers, Florian},
  journal={arXiv preprint arXiv:2601.04778},
  year={2026}
}

@inproceedings{islam2024video,
  title={Video recap: Recursive captioning of hour-long videos},
  author={Islam, Md Mohaiminul and Ho, Ngan and Yang, Xitong and Nagarajan, Tushar and Torresani, Lorenzo and Bertasius, Gedas},
  booktitle={Proceedings of the IEEE/CVF Conference on Computer Vision and Pattern Recognition},
  pages={18198--18208},
  year={2024}
}

@inproceedings{xu2024retrieval,
  title={Retrieval-augmented egocentric video captioning},
  author={Xu, Jilan and Huang, Yifei and Hou, Junlin and Chen, Guo and Zhang, Yuejie and Feng, Rui and Xie, Weidi},
  booktitle={Proceedings of the IEEE/CVF Conference on Computer Vision and Pattern Recognition},
  pages={13525--13536},
  year={2024}
}

@inproceedings{kim2024you,
  title={Do you remember? dense video captioning with cross-modal memory retrieval},
  author={Kim, Minkuk and Kim, Hyeon Bae and Moon, Jinyoung and Choi, Jinwoo and Kim, Seong Tae},
  booktitle={Proceedings of the IEEE/CVF Conference on Computer Vision and Pattern Recognition},
  pages={13894--13904},
  year={2024}
}

@article{chen2024sharegpt4video,
  title={Sharegpt4video: Improving video understanding and generation with better captions},
  author={Chen, Lin and Wei, Xilin and Li, Jinsong and Dong, Xiaoyi and Zhang, Pan and Zang, Yuhang and Chen, Zehui and Duan, Haodong and Tang, Zhenyu and Yuan, Li and others},
  journal={Advances in Neural Information Processing Systems},
  volume={37},
  pages={19472--19495},
  year={2024}
}

@inproceedings{radenovic2023filtering,
  title={Filtering, distillation, and hard negatives for vision-language pre-training},
  author={Radenovic, Filip and Dubey, Abhimanyu and Kadian, Abhishek and Mihaylov, Todor and Vandenhende, Simon and Patel, Yash and Wen, Yi and Ramanathan, Vignesh and Mahajan, Dhruv},
  booktitle={Proceedings of the IEEE/CVF conference on computer vision and pattern recognition},
  pages={6967--6977},
  year={2023}
}

@article{xie2025fg,
  title={FG-CLIP: Fine-Grained Visual and Textual Alignment},
  author={Xie, Chunyu and Wang, Bin and Kong, Fanjing and Li, Jincheng and Liang, Dawei and Zhang, Gengshen and Leng, Dawei and Yin, Yuhui},
  journal={arXiv preprint arXiv:2505.05071},
  year={2025}
}

@article{bai2025qwen2,
  title={Qwen2. 5-vl technical report},
  author={Bai, Shuai and Chen, Keqin and Liu, Xuejing and Wang, Jialin and Ge, Wenbin and Song, Sibo and Dang, Kai and Wang, Peng and Wang, Shijie and Tang, Jun and others},
  journal={arXiv preprint arXiv:2502.13923},
  year={2025}
}

@article{cho2025perceptionlm,
  title={Perceptionlm: Open-access data and models for detailed visual understanding},
  author={Cho, Jang Hyun and Madotto, Andrea and Mavroudi, Effrosyni and Afouras, Triantafyllos and Nagarajan, Tushar and Maaz, Muhammad and Song, Yale and Ma, Tengyu and Hu, Shuming and Jain, Suyog and others},
  journal={arXiv preprint arXiv:2504.13180},
  year={2025}
}

@inproceedings{chen2024panda,
  title={Panda-70m: Captioning 70m videos with multiple cross-modality teachers},
  author={Chen, Tsai-Shien and Siarohin, Aliaksandr and Menapace, Willi and Deyneka, Ekaterina and Chao, Hsiang-wei and Jeon, Byung Eun and Fang, Yuwei and Lee, Hsin-Ying and Ren, Jian and Yang, Ming-Hsuan and others},
  booktitle={Proceedings of the IEEE/CVF Conference on Computer Vision and Pattern Recognition},
  pages={13320--13331},
  year={2024}
}

@inproceedings{zhao2024distilling,
  title={Distilling vision-language models on millions of videos},
  author={Zhao, Yue and Zhao, Long and Zhou, Xingyi and Wu, Jialin and Chu, Chun-Te and Miao, Hui and Schroff, Florian and Adam, Hartwig and Liu, Ting and Gong, Boqing and others},
  booktitle={Proceedings of the IEEE/CVF Conference on Computer Vision and Pattern Recognition},
  pages={13106--13116},
  year={2024}
}

@article{Ouyang2022instructGPT,
  title={Training language models to follow instructions with human feedback},
  author={Long Ouyang and Jeff Wu and Xu Jiang and Diogo Almeida and Carroll L. Wainwright and Pamela Mishkin and Chong Zhang and Sandhini Agarwal and Katarina Slama and Alex Ray and John Schulman and Jacob Hilton and Fraser Kelton and Luke E. Miller and Maddie Simens and Amanda Askell and Peter Welinder and Paul Francis Christiano and Jan Leike and Ryan J. Lowe},
  journal={ArXiv},
  year={2022},
  volume={abs/2203.02155},
  url={https://api.semanticscholar.org/CorpusID:246426909}
}

@article{Ding2025PaMiVDPOMV,
  title={PaMi-VDPO: Mitigating Video Hallucinations by Prompt-Aware Multi-Instance Video Preference Learning},
  author={Xinpeng Ding and Kui Zhang and Jinahua Han and Lanqing Hong and Hang Xu and Xiaomeng Li},
  journal={ArXiv},
  year={2025},
  volume={abs/2504.05810},
  url={https://api.semanticscholar.org/CorpusID:277627967}
}

@article{Wang2023PaxionPA,
  title={Paxion: Patching Action Knowledge in Video-Language Foundation Models},
  author={Zhenhailong Wang and Ansel Blume and Sha Li and Genglin Liu and Jaemin Cho and Zineng Tang and Mohit Bansal and Heng Ji},
  journal={ArXiv},
  year={2023},
  volume={abs/2305.10683},
  url={https://api.semanticscholar.org/CorpusID:258762310}
}

@article{Blume2025PARTONOMYLM,
  title={PARTONOMY: Large Multimodal Models with Part-Level Visual Understanding},
  author={Ansel Blume and Jeonghwan Kim and Hyeonjeong Ha and Elen Chatikyan and Xiaomeng Jin and Khanh Duy Nguyen and Nanyun Peng and Kai-Wei Chang and Derek Hoiem and Heng Ji},
  journal={ArXiv},
  year={2025},
  volume={abs/2505.20759},
  url={https://api.semanticscholar.org/CorpusID:278911816}
}

@article{Chen2023InternVS,
  title={Intern VL: Scaling up Vision Foundation Models and Aligning for Generic Visual-Linguistic Tasks},
  author={Zhe Chen and Jiannan Wu and Wenhai Wang and Weijie Su and Guo Chen and Sen Xing and Zhong Muyan and Qinglong Zhang and Xizhou Zhu and Lewei Lu and Bin Li and Ping Luo and Tong Lu and Yu Qiao and Jifeng Dai},
  journal={2024 IEEE/CVF Conference on Computer Vision and Pattern Recognition (CVPR)},
  year={2023},
  pages={24185-24198},
  url={https://api.semanticscholar.org/CorpusID:266521410}
}

@article{Lin2023VideoLLaVALU,
  title={Video-LLaVA: Learning United Visual Representation by Alignment Before Projection},
  author={Bin Lin and Bin Zhu and Yang Ye and Munan Ning and Peng Jin and Li Yuan},
  journal={ArXiv},
  year={2023},
  volume={abs/2311.10122},
  url={https://api.semanticscholar.org/CorpusID:265281544}
}

@article{Kim2024OpenVLAAO,
  title={OpenVLA: An Open-Source Vision-Language-Action Model},
  author={Moo Jin Kim and Karl Pertsch and Siddharth Karamcheti and Ted Xiao and Ashwin Balakrishna and Suraj Nair and Rafael Rafailov and Ethan Paul Foster and Grace Lam and Pannag R. Sanketi and Quan Vuong and Thomas Kollar and Benjamin Burchfiel and Russ Tedrake and Dorsa Sadigh and Sergey Levine and Percy Liang and Chelsea Finn},
  journal={ArXiv},
  year={2024},
  volume={abs/2406.09246},
  url={https://api.semanticscholar.org/CorpusID:270440391}
}

@inproceedings{Driess2023PaLMEAE,
  title={PaLM-E: An Embodied Multimodal Language Model},
  author={Danny Driess and F. Xia and Mehdi S. M. Sajjadi and Corey Lynch and Aakanksha Chowdhery and Brian Ichter and Ayzaan Wahid and Jonathan Tompson and Quan Ho Vuong and Tianhe Yu and Wenlong Huang and Yevgen Chebotar and Pierre Sermanet and Daniel Duckworth and Sergey Levine and Vincent Vanhoucke and Karol Hausman and Marc Toussaint and Klaus Greff and Andy Zeng and Igor Mordatch and Peter R. Florence},
  booktitle={International Conference on Machine Learning},
  year={2023},
  url={https://api.semanticscholar.org/CorpusID:257364842}
}

@article{Rasheed2023GLaMMPG,
  title={GLaMM: Pixel Grounding Large Multimodal Model},
  author={Hanoona Abdul Rasheed and Muhammad Maaz and Sahal Shaji Mullappilly and Abdelrahman M. Shaker and Salman H. Khan and Hisham Cholakkal and Rao Muhammad Anwer and Eric P. Xing and Ming-Hsuan Yang and Fahad Shahbaz Khan},
  journal={2024 IEEE/CVF Conference on Computer Vision and Pattern Recognition (CVPR)},
  year={2023},
  pages={13009-13018},
  url={https://api.semanticscholar.org/CorpusID:265043538}
}

@article{Xu2016MSRVTTAL,
  title={MSR-VTT: A Large Video Description Dataset for Bridging Video and Language},
  author={Jun Xu and Tao Mei and Ting Yao and Yong Rui},
  journal={2016 IEEE Conference on Computer Vision and Pattern Recognition (CVPR)},
  year={2016},
  pages={5288-5296},
  url={https://api.semanticscholar.org/CorpusID:206594535}
}

@article{Rawal2025ARGUSHA,
  title={ARGUS: Hallucination and Omission Evaluation in Video-LLMs},
  author={Ruchit Rawal and Reza Shirkavand and Heng Huang and Gowthami Somepalli and Tom Goldstein},
  journal={ArXiv},
  year={2025},
  volume={abs/2506.07371},
  url={https://api.semanticscholar.org/CorpusID:279251247}
}

@article{Bolya2025PerceptionET,
  title={Perception Encoder: The best visual embeddings are not at the output of the network},
  author={Daniel Bolya and Po-Yao Huang and Peize Sun and Jang Hyun Cho and Andrea Madotto and Chen Wei and Tengyu Ma and Jiale Zhi and Jathushan Rajasegaran and Hanoona Abdul Rasheed and Junke Wang and Marco Monteiro and Hu Xu and Shiyu Dong and Nikhila Ravi and Shang-Wen Li and Piotr Doll'ar and Christoph Feichtenhofer},
  journal={ArXiv},
  year={2025},
  volume={abs/2504.13181},
  url={https://api.semanticscholar.org/CorpusID:277856792}
}

@article{Chai2024AuroraCapEP,
  title={AuroraCap: Efficient, Performant Video Detailed Captioning and a New Benchmark},
  author={Wenhao Chai and Enxin Song and Yilun Du and Chenlin Meng and Vashisht Madhavan and Omer Bar-Tal and Jeng Neng Hwang and Saining Xie and Christopher D. Manning},
  journal={ArXiv},
  year={2024},
  volume={abs/2410.03051},
  url={https://api.semanticscholar.org/CorpusID:273162259}
}

@inproceedings{Radford2021clip,
  title={Learning Transferable Visual Models From Natural Language Supervision},
  author={Alec Radford and Jong Wook Kim and Chris Hallacy and Aditya Ramesh and Gabriel Goh and Sandhini Agarwal and Girish Sastry and Amanda Askell and Pamela Mishkin and Jack Clark and Gretchen Krueger and Ilya Sutskever},
  booktitle={International Conference on Machine Learning},
  year={2021},
  url={https://api.semanticscholar.org/CorpusID:231591445}
}

@article{Matsuda2025vela,
  title={VELA: An LLM-Hybrid-as-a-Judge Approach for Evaluating Long Image Captions},
  author={Kazuki Matsuda and Yuiga Wada and Shinnosuke Hirano and Seitaro Otsuki and Komei Sugiura},
  journal={ArXiv},
  year={2025},
  volume={abs/2509.25818},
  url={https://api.semanticscholar.org/CorpusID:281681513}
}

@article{Gu2024ASO,
  title={A Survey on LLM-as-a-Judge},
  author={Jiawei Gu and Xuhui Jiang and Zhichao Shi and Hexiang Tan and Xuehao Zhai and Chengjin Xu and Wei Li and Yinghan Shen and Shengjie Ma and Honghao Liu and Yuanzhuo Wang and Jian Guo},
  journal={ArXiv},
  year={2024},
  volume={abs/2411.15594},
  url={https://api.semanticscholar.org/CorpusID:274234014}
}

@inproceedings{Loshchilov2017adamw,
  title={Decoupled Weight Decay Regularization},
  author={Ilya Loshchilov and Frank Hutter},
  booktitle={International Conference on Learning Representations},
  year={2017},
  url={https://api.semanticscholar.org/CorpusID:53592270}
}

@article{dubey2024llama3,
  title={The llama 3 herd of models},
  author={Dubey, Abhimanyu and Jauhri, Abhinav and Pandey, Abhinav and Kadian, Abhishek and Al-Dahle, Ahmad and Letman, Aiesha and Mathur, Akhil and Schelten, Alan and Yang, Amy and Fan, Angela and others},
  journal={arXiv e-prints},
  pages={arXiv--2407},
  year={2024}
}

@article{pal2024dpop,
  title={Smaug: Fixing failure modes of preference optimisation with dpo-positive},
  author={Pal, Arka and Karkhanis, Deep and Dooley, Samuel and Roberts, Manley and Naidu, Siddartha and White, Colin},
  journal={arXiv preprint arXiv:2402.13228},
  year={2024}
}

@online{IntroducingGPTOSS2025,
  title = {Introducing {{GPT-OSS}}},
  date = {2025-08-05},
  url = {https://openai.com/index/introducing-gpt-oss/},
  urldate = {2025-08-05},
  langid = {american}
}

@inproceedings{goyal2017ssv2,
  title={The" something something" video database for learning and evaluating visual common sense},
  author={Goyal, Raghav and Ebrahimi Kahou, Samira and Michalski, Vincent and Materzynska, Joanna and Westphal, Susanne and Kim, Heuna and Haenel, Valentin and Fruend, Ingo and Yianilos, Peter and Mueller-Freitag, Moritz and others},
  booktitle={Proceedings of the IEEE international conference on computer vision},
  pages={5842--5850},
  year={2017}
}

@article{Hu2021lora,
  title={LoRA: Low-Rank Adaptation of Large Language Models},
  author={J. Edward Hu and Yelong Shen and Phillip Wallis and Zeyuan Allen-Zhu and Yuanzhi Li and Shean Wang and Weizhu Chen},
  journal={ArXiv},
  year={2021},
  volume={abs/2106.09685},
  url={https://api.semanticscholar.org/CorpusID:235458009}
}

@inproceedings{Kim2024Finer,
  title={Finer: Investigating and Enhancing Fine-Grained Visual Concept Recognition in Large Vision Language Models},
  author={Jeonghwan Kim and Heng Ji},
  booktitle={Conference on Empirical Methods in Natural Language Processing},
  year={2024},
  url={https://api.semanticscholar.org/CorpusID:267938790}
}

@article{Liu2023Lostinthemiddle,
  title={Lost in the Middle: How Language Models Use Long Contexts},
  author={Nelson F. Liu and Kevin Lin and John Hewitt and Ashwin Paranjape and Michele Bevilacqua and Fabio Petroni and Percy Liang},
  journal={Transactions of the Association for Computational Linguistics},
  year={2023},
  volume={12},
  pages={157-173},
  url={https://api.semanticscholar.org/CorpusID:259360665}
}

@inproceedings{yang2025onpolicydpo,
  title={Mitigating hallucinations in large vision-language models via dpo: On-policy data hold the key},
  author={Yang, Zhihe and Luo, Xufang and Han, Dongqi and Xu, Yunjian and Li, Dongsheng},
  booktitle={Proceedings of the Computer Vision and Pattern Recognition Conference},
  pages={10610--10620},
  year={2025}
}

@article{shenfeld2025rlsrazor,
  title={RL's Razor: Why Online Reinforcement Learning Forgets Less},
  author={Shenfeld, Idan and Pari, Jyothish and Agrawal, Pulkit},
  journal={arXiv preprint arXiv:2509.04259},
  year={2025}
}

@inproceedings{yu2025rlaifv,
  title={Rlaif-v: Open-source ai feedback leads to super gpt-4v trustworthiness},
  author={Yu, Tianyu and Zhang, Haoye and Li, Qiming and Xu, Qixin and Yao, Yuan and Chen, Da and Lu, Xiaoman and Cui, Ganqu and Dang, Yunkai and He, Taiwen and others},
  booktitle={Proceedings of the Computer Vision and Pattern Recognition Conference},
  pages={19985--19995},
  year={2025}
}

@inproceedings{zhai2023siglip,
  title={Sigmoid loss for language image pre-training},
  author={Zhai, Xiaohua and Mustafa, Basil and Kolesnikov, Alexander and Beyer, Lucas},
  booktitle={Proceedings of the IEEE/CVF international conference on computer vision},
  pages={11975--11986},
  year={2023}
}

@inproceedings{jia2021align,
  title={Scaling up visual and vision-language representation learning with noisy text supervision},
  author={Jia, Chao and Yang, Yinfei and Xia, Ye and Chen, Yi-Ting and Parekh, Zarana and Pham, Hieu and Le, Quoc and Sung, Yun-Hsuan and Li, Zhen and Duerig, Tom},
  booktitle={International conference on machine learning},
  pages={4904--4916},
  year={2021},
  organization={PMLR}
}

@article{gu2024llm_as_judge_survey,
  title={A survey on llm-as-a-judge},
  author={Gu, Jiawei and Jiang, Xuhui and Shi, Zhichao and Tan, Hexiang and Zhai, Xuehao and Xu, Chengjin and Li, Wei and Shen, Yinghan and Ma, Shengjie and Liu, Honghao and others},
  journal={The Innovation},
  year={2024},
  publisher={Elsevier}
}

@inproceedings{burges2005ranknet,
  title={Learning to rank using gradient descent},
  author={Burges, Chris and Shaked, Tal and Renshaw, Erin and Lazier, Ari and Deeds, Matt and Hamilton, Nicole and Hullender, Greg},
  booktitle={Proceedings of the 22nd international conference on Machine learning},
  pages={89--96},
  year={2005}
}
